\documentclass[twocolumn]{article}
\usepackage[utf8]{inputenc}
\usepackage[english]{babel}
\usepackage{amsmath}
\usepackage{amsthm}
\usepackage{graphicx}
\usepackage{xcolor}
\usepackage{amssymb}
\usepackage{floatrow}
\usepackage{multirow}
\usepackage{tikz}

\usepackage{color,soul}

\usepackage{authblk}

\usepackage[ruled,vlined, linesnumbered]{algorithm2e}

\let\oldnl\nl
\newcommand{\nonl}{\renewcommand{\nl}{\let\nl\oldnl}}

\newcommand{\dimx}{\boldsymbol{N}}
\newcommand{\nll}{\boldsymbol{\ell}}

\usepackage{lstautogobble}

\usepackage{hyperref}
\hypersetup{
     colorlinks = true,
     linkcolor = green!60!black,
     anchorcolor = green!50!black,
     citecolor = green!60!black,
     filecolor = green!60!black,
     urlcolor = green!60!black
     }
\usepackage[nameinlink,noabbrev]{cleveref}
\usepackage[a4paper, total={7in, 10in}]{geometry}

\newtheorem{theorem}{Theorem}

\newtheorem{proposition}[theorem]{Proposition}

\newenvironment{definition}[1][Definition]{\begin{trivlist}
\item[\hskip \labelsep {\bfseries #1}]}{\end{trivlist}}

\definecolor{mypurple}{RGB}{153, 50, 204}

\newcommand{\comment}[1]{}

\title{\textbf{Latent Space Data Assimilation by using Deep Learning 
}\medskip}

\author[1,2,4]{Mathis Peyron}
\author[5,1]{Anthony Fillion}
\author[4,1]{Selime G\"urol}
\author[1]{Victor Marchais}
\author[5,1]{Serge Gratton}
\author[3,1]{Pierre Boudier}
\author[2]{Gael Goret}
\medskip

\affil[1]{ANITI, Universit\'e de Toulouse}
\affil[2]{Atos BDS R\&D AI4Sim}
\affil[3]{NVIDIA}
\affil[4]{CERFACS}
\affil[5]{UFTMIP}

\date{}
\begin{document}

\maketitle

\begin{abstract}
Performing Data Assimilation (DA) at a low cost is of prime concern in Earth system modeling, particularly at the time of big data where huge quantities of observations are available. Capitalizing on the ability of Neural Networks techniques for approximating the solution of PDE's, we incorporate Deep Learning (DL) methods into a DA framework. More precisely, we exploit the latent structure provided by autoencoders (AEs) to design an Ensemble Transform Kalman Filter with model error (ETKF-Q) in the latent space. Model dynamics are also propagated within the latent space via a surrogate neural network. 
This novel {ETKF-Q-Latent (thereafter referred to as ETKF-Q-L)} algorithm is tested on a tailored instructional version of Lorenz 96 equations, named \textit{the augmented Lorenz 96 system}: it possesses a latent structure that accurately represents the observed dynamics. Numerical experiments based on this particular system evidence that the {ETKF-Q-L} approach both reduces the computational cost and provides better accuracy than state of the art algorithms, such as the ETKF-Q.

\end{abstract}

\section{Introduction}

    Data Assimilation estimates the state of a system $\boldsymbol{x} \in \mathbb{R}^n$ given two sources of information: a model that provides a background knowledge $\boldsymbol{x}^b \in \mathbb{R}^n$ and an observation vector $\boldsymbol{y} \in \mathbb{R}^p$. We can either observe the system directly - partially or entirely - or implicitly through an observation operator $\mathcal{H}$, yielding the observation relation $\boldsymbol{y} = \mathcal{H}(\boldsymbol{x})$~\cite{BocquetSiam, BocquetLecture}.
   
    Most variational and ensemble DA algorithms encounter two problems that will be considered in the present paper.

    First, they are largely grounded on the Kalman filter which assumes that errors follow a Gaussian distribution and the physical model is mildly nonlinear. Another issue is the cost of applying the propagation model that can be computationally prohibitive. Thus, we propose to follow a data-driven approach which can efficiently represent nonlinear dynamics as long as a sufficient amount of data is given. They can also make better use of big data and exploit better the future computing hardware.\\
    
    In the last decade, machine learning (ML) techniques managed to outperform existing methods in image classification \cite{Krizhevsky, le2012building, he2015deep, simonyan2015deep}, segmentation \cite{qi2017pointnet, garcia2017review}, Natural Language Processing (NLP) \cite{otter2019survey, gardner2018allennlp}, language translation \cite{sutskever2014sequence, wu2016googles} and in mastering go game \cite{silver2016mastering, silver2017mastering} for instance. 
    It is now broadening to new areas like computational physics where it achieves interesting results too, as exemplified by the so-called physically informed artificial neural networks \cite{raissi2017physics1, raissi2017physics2, mohan2020embedding}.
    
    Nevertheless, the widespread applicability of ML to plethora of problems is hampered by significant concerns like the difficulty to incorporate physical knowledge in DL frameworks \cite{vonrueden2020informed, Boukabara2020} or the existence of adversarial examples \cite{Szegedy2013IntriguingPO, CarliniWagner, rottmann2020detection} that reduces practical applications in sensitive fields (e.g. automated driving). We can also mention the absence of uncertainty estimation analysis \cite{Reichstein2019DeepLA}.\\
    
    As suggested by \cite{ECMWFarticle}, Data Assimilation and Deep Learning can take advantage of each other as they are complementary and have also similarities.  
    Therefore, coupling DA and DL is quite a natural approach.
    In this study, we are interested in utilizing DL tools within a DA framework to:

    \begin{itemize}
        \item reduce the computational cost and memory storage by performing calculations in a reduced space (latent space).
        \item get a better accuracy by exploiting the latent structure and a surrogate network.
        \item propose a novel and promising framework that could be extended to other DA methods and NNs. The proposed method simplicity is also a great advantage: there is no need to change the DA algorithm, only the operators.
    \end{itemize}

    \subsection{Related work and our contributions}
        \label{section:Related work}
        Optimizing DA methods is a key issue in Earth system modeling, particularly in the context of big data where huge quantities of observations are now available (thanks to remote sensing among other \cite{Kuenzer}), and where the increase of the complexity of physical models goes hand in hand with the augmentation of the computational resources needed in applications. For this reason, model reduction techniques and surrogate models have been investigated by the DA community.\\
        
        As shown in \cite[Chapter 5]{BocquetSiam}, DA reduction techniques rely on multiple mathematical tools in order to alleviate the computational cost: singular value decompositions, principal component analysis (PCA) or proper orthogonal decomposition (POD), spectral decomposition (\textit{e.g.} Fourier series), wavelets and curvelets for instance. 
        Reduction methods in DA can be divided into two groups: either they aim at reducing the numerical cost of the time integration of the model or they improve the DA method itself.
        
        Probably the most direct approach in the first category consists in simplifying the model (see \cite{Miller1989}), whereas more sophisticated approaches rely on POD \cite{Cao2006b, artana} or on wavelets \cite{TANGBORN2000}. Rank reduction techniques have been investigated as well for both variational and sequential methods: the core change is to introduce low rank approximations of the covariance matrices, without changing the complexity of the time integration. In this case, some methods focus on factorizing the model error covariance matrix by possibly relying on sparse structures (see \cite{Bannister2008} for a review). Propagating such a matrix through time is also expensive, hence the idea is to use a Truncated Singular Value Decomposition (TSVD) on the model linear operator \cite{Cohn1996}. 
        
        When coming to improving the DA algorithm by space reduction, we find the Singular Evolutive Extended Kalman Filter (SEEK) \cite{TUANPHAM1998} and, using sampling techniques, ensemble DA methods like Ensemble Kalman Filter (EnKF), the reduced-rank square root filter (RRSQRT), the stochastic EnKF and the deterministic EnKF.\\
        
        Deep Learning also proposes reduction techniques along with surrogate models. It is also establishing itself in computational physics. Recent research shows that PDEs can be solved efficiently by using DL methods and the computational cost of simulations can be reduced by considering the latent space structure provided by NNs \cite{raissi2017physics1, raissi2017physics2,li2020fourier,seo2019graphs, Han_2018, Khoo2018SolvingPP}. For instance, the Physics Informed Neural Networks (PINNs) \cite{raissi2017physics1, raissi2017physics2} are capable to properly learn the solution of a nonlinear PDE while being computationally cheap. Also, one of the most promising research work, though not investigated here, consists in making the most of Fourier transforms to solve PDEs \cite{li2020fourier}: this technique has the advantage to be both accurate and computationally efficient. 
        
        Creating a latent space is mostly performed with well-known networks like autoencoders (AE), convolutional autoencoders (CAE) or variational autoencoders (VAE). The aim in creating a latent space is either to get a better accuracy or to reduce the computational cost. For example, in \cite{Canchumuni2019} a convolutional VAE is used to construct a continuous parameterization for facies in order to preserve the geological realism of the model (\textit{i.e.} predictions for oil and gas reservoirs). This leads to a better accuracy.
        In \cite{Wiewel2018, Wiewel2020}, the authors are mainly interested in computational speed up and  {robust long-term predictions} for fluid flows simulations. 
        They therefore demonstrate the capability of data-driven approaches for modeling fluid dynamics: they implemented a CAE for spatial compression and stacked long-short term memory (LSTM) layers to define their surrogate network, \textit{i.e.} a network that performs time propagation. They achieve a compression ratio of $256$ enabling them to replace heavy fluid flows simulations. Similarly, \cite{maulik2020reducedorder} resorts to a CAE for space reduction and recurrent neural networks for the time propagation. About \cite{Mack2020AttentionbasedCA}, in addition to Control Variable Transform (CVT) they take advantage of a CAE inspired by the image compression field that uses state-of-the-art deep learning techniques. In \cite{Raissi_2018}, fully-connected layers are considered to perform the reduction.
        Lastly, we can quote \cite{FultonDeformable2018} where the authors utilize an AE to encode 3D virtual figurines, then deform their motions within the latent space and finally get the new pose in the full space. 
        Time propagation within the latent space is performed with a mere matrix product given by the so-called Koopman operator. The reduced space obtained with this method is of much smaller dimension than linear model reduction techniques, therefore it yields faster and more accurate results along with improved robustness.\\
    
        When including model reduction techniques within DA frameworks, we find \cite{Canchumuni2019} and \cite{Mack2020AttentionbasedCA} which both employ neural networks in their data assimilation architectures. Yet, they do not have the same purposes. Indeed, \cite{Canchumuni2019} trains a {convolutional} variational autoencoder (CVAE) with the aim to generate realistic geological facies {by introducing a new continuous parameterization} which was not properly achieved with prior techniques. 
        They use a type of ensemble smoother based on multiple data assimilation and assimilate the data in the new parametrization (latent) space which leads to a better accuracy. They do not use a surrogate network in their DA method.
        
        Regarding \cite{Mack2020AttentionbasedCA}, the goal is to reduce the physical domain into a latent space to speed up computations in the context of an ensemble based DA without temporal aspect.
        Besides the temporal aspect, the main difference with our methodology is that they introduce an observation encoder network which includes an interpolation operator to map the observations to the full space. This observation network maps the observations to the latent space where data assimilation is performed. Having an interpolation operator may introduce additional errors during DA. Whereas in our method, observations stay in their original space and we do not need to apply any transformation. In their latent space DA approach, they also only focus on the computational cost gain.
        
        As for surrogate networks coupled with DA algorithms, we have \cite{pawar2020data} and \cite{Brajard_2020} which both consider Lorenz 96 system. In \cite{pawar2020data}, Pawar  and San model unresolved flow dynamics with a surrogate network and learn the correlation between resolved flow processes and unresolved subgrid variables thanks to a set of NNs. Whereas we both aim to more accurately forecast, their approach does not involve any space reduction technique. {Their main motivation is not  on reducing the computational cost either}. Regarding \cite{Brajard_2020}, Brajard et al. use an ensemble Kalman filter algorithm combined to a convolutional neural network with \textit{skipconnections} \cite{he2015deep}. What is remarkable here is that the surrogate model is iteratively trained with the data assimilation algorithm, whereas in our case NNs are not aware of the existence of an outer data assimilation process: our networks are trained independently of the ETKF-Q algorithm. Nonetheless, Brajard et al. consider neither standard space reduction techniques nor latent spaces obtained with NNs like we do.\\

        A first ingredient in the approach we introduce in the present paper is to replace the (supposedly expensive) time integration of the model with a NN surrogate. Time stepping methods based on surrogates have already been explored by several authors \cite{Lu2007SequentialDA, Wiewel2018,Wiewel2020,maulik2020reducedorder,Vlachas2018,Brajard_2020,pawar2020data}.
        A key question that comes immediately is to ensure the time stability of the resulting scheme~(see~\cite{Haber_2017,haber2019imexnet} for an investigation within deep learning surrogate models) when the model is repeatedly called to propagate the state over several time steps. 
        To this aim, we introduce in the training loss an explicit stabilization that involves a  penalization of the growth of model iterations.\\
        
        We  also  address  the  question  of incorporating latent spaces  in  DA  frameworks. In  general,  DA  and  ML  are not really coupled, model reduction or surrogate models are used only for the time propagation within an existing DA algorithm. We believe our approach is original in that it performs DA directly in the NN latent space. This approach is appealing, since it fully exploits an underlying geometry and leads to perform the computations mostly in the space where they are cheap (the latent space).  We shall see that to obtain good results, a special care has to be taken in the description of the dynamical and observational errors.\\
        
        Therefore, in this paper we propose a new latent space DA methodology, that is the ETKF-Q-L one, which:
        
        \begin{enumerate}
            \item explores the ability of DL for creating a $\nll$-dimensional reduced space, based on the assumption that a latent space of size $\nll$ which accurately represents the full dynamics exists. This is achieved with an autoencoder.
            
            \item defines a surrogate network within the latent space that performs time propagation. An innovative iterative training enforces the surrogate to be stable over time.
            
            \item implements an ensemble DA algorithm within the learned $\nll$-dimensional latent space thanks to the AE and the surrogate network.
        \end{enumerate}
        
        Interestingly, $1.$ and $2.$ can be performed in an all-at-once approach by training both the AE and the surrogate at the same time through a well-suited custom loss function. The training set is an ensemble of simulations of the physical system which lies in $\mathbb{R}^{\dimx}$.
        For short, the proposed methodology provides the following advantages:
        \begin{enumerate}
            \item since any DA algorithm requires to store vectors lying in the model space, discovering a lower dimensional representation induces a reduction in memory needs and computational cost.
        
            \item performing the DA linear analysis in the latent space obtained by AE is less susceptible to yield non-physical solutions since the decoder is a nonlinear transformation that fits the manifold where state trajectory statistically belongs, when such a structure does exist.
        \end{enumerate}
        
        We show the relevance of our approach on a $400$-dimensional system possessing an underlying dynamics that follows the $40$-variables Lorenz 96 equations. The existence of this latent physics is ensured by construction of the full space dynamics which is built upon Lorenz 96 system.\\

    \subsection{Organization of this work}

        The remainder of this work is organized as follows: in \cref{section:methodology1}, we precisely detail the role of the AE  NN structure for the latent space computation. We also present the surrogate network used for time stepping  and expose the loss function that allows the joint learning of these two networks.
    
        In \cref{sec:DA}, we first remind some general DA concepts, then we explain an ensemble algorithm that takes model error into account that is the ETKF-Q algorithm. Lastly, we detail our latent DA framework namely the ETKF-Q-L method, based on the ETKF-Q algorithm associated with the AE and the surrogate. 
    
        In \cref{section:numerical experiments}, we present the experimental context chosen to benchmark our approach, \textit{i.e.} the augmented Lorenz 96 dynamics, the NN architectures and the training setting along with the DA context. We show the performance of the new DA method on the augmented Lorenz 96 system. A grid search algorithm is used to tune the DA algorithm parameters.

\section{Latent Space Dynamics}
\label{section:methodology1}

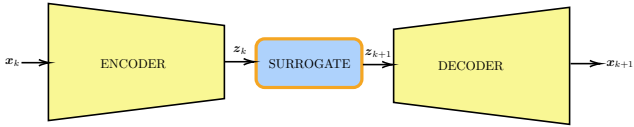
\begin{figure}[t]
            \centering
                \hspace*{-0.0cm}\scalebox{0.8}{
                    \tikzset{every picture/.style={line width=0.75pt}} 
                    \begin{tikzpicture}[x=0.75pt,y=0.75pt,yscale=-0.6,xscale=0.6, every node/.style={scale=0.6}]
                    
                    \draw  [color={rgb, 255:red, 245; green, 166; blue, 35 }  ,draw opacity=1 ][fill={rgb, 255:red, 174; green, 210; blue, 252 }  ,fill opacity=1 ][line width=1.5]  (265.3,89.32) .. controls (265.3,84.4) and (269.3,80.4) .. (274.22,80.4) -- (366.37,80.4) .. controls (371.3,80.4) and (375.3,84.4) .. (375.3,89.32) -- (375.3,122.47) .. controls (375.3,127.4) and (371.3,131.4) .. (366.37,131.4) -- (274.22,131.4) .. controls (269.3,131.4) and (265.3,127.4) .. (265.3,122.47) -- cycle ;
                    \draw  [fill={rgb, 255:red, 249; green, 247; blue, 148 }  ,fill opacity=1 ] (232.3,143.05) -- (49.3,166.05) -- (49.3,43.4) -- (232.3,66.4) -- cycle ;
                    \draw  [fill={rgb, 255:red, 249; green, 247; blue, 148 }  ,fill opacity=1 ] (408.3,70.4) -- (591.3,47.4) -- (591.3,170.05) -- (408.3,147.05) -- cycle ;
                    \draw    (21.3,105.4) -- (48.3,105.4) ;
                    \draw [shift={(50.3,105.4)}, rotate = 180] [color={rgb, 255:red, 0; green, 0; blue, 0 }  ][line width=0.75]    (10.93,-3.29) .. controls (6.95,-1.4) and (3.31,-0.3) .. (0,0) .. controls (3.31,0.3) and (6.95,1.4) .. (10.93,3.29)   ;
                    \draw    (232.3,104.4) -- (261.3,104.4) ;
                    \draw [shift={(263.3,104.4)}, rotate = 180] [color={rgb, 255:red, 0; green, 0; blue, 0 }  ][line width=0.75]    (10.93,-3.29) .. controls (6.95,-1.4) and (3.31,-0.3) .. (0,0) .. controls (3.31,0.3) and (6.95,1.4) .. (10.93,3.29)   ;
                    \draw    (376.3,107.4) -- (405.3,107.4) ;
                    \draw [shift={(407.3,107.4)}, rotate = 180] [color={rgb, 255:red, 0; green, 0; blue, 0 }  ][line width=0.75]    (10.93,-3.29) .. controls (6.95,-1.4) and (3.31,-0.3) .. (0,0) .. controls (3.31,0.3) and (6.95,1.4) .. (10.93,3.29)   ;
                    \draw    (592.3,106.4) -- (621.3,106.4) ;
                    \draw [shift={(623.3,106.4)}, rotate = 180] [color={rgb, 255:red, 0; green, 0; blue, 0 }  ][line width=0.75]    (10.93,-3.29) .. controls (6.95,-1.4) and (3.31,-0.3) .. (0,0) .. controls (3.31,0.3) and (6.95,1.4) .. (10.93,3.29)   ;
                    
                    \draw (277,100) node [anchor=north west][inner sep=0.75pt] [align=left] {SURROGATE};
                    \draw (102,100) node [anchor=north west][inner sep=0.75pt]   [align=left] {ENCODER};
                    \draw (453,102) node [anchor=north west][inner sep=0.75pt]   [align=left] {DECODER};
                    \draw (628,101) node [anchor=north west][inner sep=0.75pt]   [align=left] {$\boldsymbol{x}_{k+1}$};
                    \draw (3,99) node [anchor=north west][inner sep=0.75pt]   [align=left] {$\boldsymbol{x}_k$};
                    \draw (240,87) node [anchor=north west][inner sep=0.75pt]   [align=left] {$\boldsymbol{z}_k$};
                    \draw (377,90) node [anchor=north west][inner sep=0.75pt]   [align=left] {$\boldsymbol{z}_{k+1}$};
                    
                    \end{tikzpicture}
                }
            \caption{Training architecture of the autoencoder and the surrogate: the encoder maps from $\mathbb{R}^{\dimx}$ to $\mathbb{R}^{\nll}$ and the decoder performs the reverse operation. The surrogate, \textit{i.e.} the time propagator operates within the latent space. Both networks are trained together.}
            \label{fig:TrainingArchitecture}
        \end{figure}

    In this section we explain how to approximate the model dynamics under consideration in a latent space. We first briefly present the autoencoder (AE) structures, which are widely used for dimension reduction \cite{Kramer1991}, and then provide more details on the surrogate network which is nested in the AE (\cref{fig:TrainingArchitecture}). 
    We also discuss how to obtain a stable trajectory of the dynamics by reformulating the loss function.\\
    
    Resorting to a latent space data assimilation is also motivated by \cite{TrevisanUboldi2004, CarrassiTrevisan2007, Trevisan2010, Trevisan2011}. In the case of nonlinear chaotic dynamics, the state space can be indeed divided into unstable and stable subspaces. It turns out that perturbations mostly live in the unstable manifold which is in general of much smaller dimension than the full space. In \cite{bocquet:hal-01592362}, these results are extended to the Kalman smoother and ensemble formulations of the Kalman filter and smoother are considered.
    This motivates the link we make between latent DA and unstable manifold
    theory; our goal is to learn a latent space that would capture the unstable manifold. Indeed, as shown by \cite{UboldiTrevisan2006,CarrassiTrevisan2008}, tracking the unstable directions is of primary relevance to properly perform the assimilation in possibly high-dimensional models such as the atmospheric or oceanic models.

    \subsection{Autoencoders}
        \label{sec:AE}
    
        Autoencoders are a type of NNs that are trained to reproduce the input data by enforcing them to be accurately represented in a lower dimension \cite{Kramer1991}. They consist of an encoder and a decoder trained together.
        Given data lying in $\mathbb{R}^{\dimx}$, the encoder maps from $\mathbb{R}^{\dimx}$ to $\mathbb{R}^{\nll}$ (with $\nll<<\dimx$) and is generally made of successive fully connected or convolutional layers with decreasing dimensions. The decoder performs the reverse operation and therefore mirrors encoder's layers. A common loss function is the Mean Square Error (MSE):
        \begin{align}
        \label{eq:AEloss}
            \small{
            \text{MSE}\left(\boldsymbol{x},\mathcal{D}\left(\mathcal{E}\left(\boldsymbol{x}\right)\right)\right) = \frac{1}{\dimx} \sum_{i=1}^{\dimx} \left(\boldsymbol{x}^{(i)} - \left[\mathcal{D}\left(\mathcal{E}\left(\boldsymbol{x}\right)\right)\right]^{(i)}\right)^2
            }
        \end{align}
        where $\boldsymbol{x} \in \mathbb{R}^{\dimx}$, $\boldsymbol{x}^{(i)}$ denotes the $i$-th element of vector $\boldsymbol{x}$, $\mathcal{E}$ and $\mathcal{D}$ denote the encoder and the decoder, respectively. Note that unlike Principal Component Analysis (PCA), AEs leverage nonlinear transformations and are thus better suited to handling nonlinearities \cite{Kramer1991}.\\
        
        In order to apprehend how AEs work, let us consider MNIST dataset\footnote{see \url{http://yann.lecun.com/exdb/mnist/} for more details} made of hand-written digits stored as $28 \times 28$ images or vectors of size $784$: a very simple encoder could contain $4$ fully connected layers whose input dimensions could be as follows: $784, 300, 150, 100, 20$. Then, the decoder would hold the same number of layers in the reverse order (\textit{i.e.} with dimensions $20, 100, 150, 300, 784$, respectively). Hence the bottleneck structure of AE networks with the reduced space at the encoder-decoder junction point.\\
    
        While learning, encoder and decoder's weights are modified so that the autoencoder can reconstruct the input digit with the strong requirement that data have to be well represented in a $\nll$-dimensional space called the \textit{latent space}.

        AE's quality is highly impacted by the type and the number of layers, their dimensions and the size of the latent space chosen. The major issue, that is used tackled numerically by hyperparameter tuning, is to be able to find the smallest latent space that enables to represent the data the more accurately as possible.\\
        
        In this study, which is more of a proof of concept, we assume available a system of size $\dimx$ for which a latent space of lower dimension $\nll$ is deemed to exist and in which the observed dynamical system can be described. Our encoder maps from $\mathbb{R}^{\dimx}$ to $\mathbb{R}^{\nll}$ and the decoder performs the reverse operation. We want to emphasize that there is absolutely no reason for the latent space of dimension $\nll$ produced by the autoencoder to be unique. Indeed, the loss function used in the training promotes a coherence between the triplet consisting of the decoder, the encoder and the latent space on the one hand, and data on the other hand. Whenever one particular latent space is discovered (the network is completely free in the way it designs the latent space), other latent spaces exist as well, obtained by transformations such as rotations, or changes of scales.
   
    \subsection{Surrogate network and stability}
        We want to train a surrogate network such that time propagation of the model dynamics can be performed in the latent space (obtained by the AE).
        Therefore, our surrogate network is estimated using encoded data and outputs a transformation acting on, and producing latent vectors. Just like the AE, a first idea would consist in training the surrogate network through a MSE loss function as follows:
        \begin{equation}
            \label{eq:reglossSurrogate}
            \text{MSE}\left(\boldsymbol{x}_{k+1},\mathcal{T}\left(\boldsymbol{x}_k\right)\right) = \frac{1}{\dimx} \sum_{i=1}^{\dimx} \left(\boldsymbol{x}_{k+1}^{(i)} - \left[\mathcal{T}\left(\boldsymbol{x}_k\right)\right]^{(i)}\right)^2
        \end{equation}
        where $\boldsymbol{x}_k, \boldsymbol{x}_{k+1} \in \mathbb{R}^{\dimx}$ are the state vectors at time $t_k$ and $t_{k+1}$, respectively and operator $\mathcal{T}$ is such that $\mathcal{T}\left(.\right) = \mathcal{D}\left(\mathcal{S}\left(\mathcal{E}\left(.\right)\right)\right)$ with $\mathcal{E}$, $\mathcal{S}$ and $\mathcal{D}$ denoting the encoder, the surrogate and the decoder, respectively.\\
        
        Nonetheless, training our surrogate with this loss function (\cref{eq:reglossSurrogate}) does not yield a stable solution. This is especially easy to understand when the dynamics under consideration is chaotic, as often the case in data assimilation. In this case, if the non-vanishing components of the dynamics are not represented with enough accuracy, the surrogate dynamics is expected to be of insufficient quality. This is even worse in the case where the original dynamics would exhibit conservative components; if the surrogate dynamics does not capture these components accurately enough, it is easy to conceive that nonphysical unstable subspaces may occur, making the latent space time stepping with the surrogate inappropriate for DA.\\
         
        Issues related to stable NNs approximation of time stepping methods have already been investigated in the literature though outside of our DA context.
       
        They have been linked to exploding or vanishing gradients issues and NNs' robustness as well. \cite{Haber_2017, haber2019imexnet} get some insights in this direction by proposing groundbreaking methods to make deep neural networks stable. However, the problem they address is not exactly the one we are looking at: they focus on Deep Neural Networks' (DNNs) robustness to input perturbation, on their capability to distinguish between two initial vector states, \textit{i.e.} not to bring both of them to $0$ nor making them diverging. 
        
        Within the framework of DA, the presence of non-physical unstable dynamics components is controlled by using a simple penalty approach involving a technique we describe now.
    
        Our method relies on a chained loss function, meaning that we train the surrogate to predict $c$ successive states to enforce stability. In practice, given $\boldsymbol{x}_k \in \mathbb{R}^{\dimx}$, the encoder yields $\boldsymbol{z}_k \in \mathbb{R}^{\nll}$. Then, the surrogate outputs $\boldsymbol{z}_{k+1}, \dots \boldsymbol{z}_{k+c}$ which are all decoded afterwards and their distances to the ground truth states are measured through a custom loss function defined as follows:
        \begin{align}
            \label{eq:iterloss}
            \frac{1}{C} \sum_{c=1}^{C} \text{MSE}\left(\mathcal{T}^c\left(\boldsymbol{x}_k\right), \boldsymbol{x}_{k+c}\right)
        \end{align}
        where $\mathcal{T}^c$ is a straightforward extension of operator $\mathcal{T}$: $\mathcal{T}^c\left(.\right) = \mathcal{D}\left(\mathcal{S}^c\left(\mathcal{E}\right)\right)$. Regarding $\mathcal{S}^c$, it means that the surrogate is applied $c$ times in a row over the given data.\\
          
        One remaining question is the number of iterations $C$ we need to perform in order to achieve this stability criterion: according to our numerical experiments based on the augmented Lorenz 96 system, just $2$ consecutive predictions already guarantee a stable behaviour. In the numerical tests, we pick this parameter in $\{2,3,4\}$.\\

      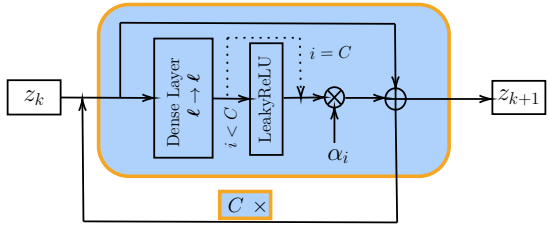
\begin{figure}[t]
    \centering
        \scalebox{0.9}{
            \tikzset{every picture/.style={line width=0.75pt}} 

            \begin{tikzpicture}[x=0.75pt,y=0.75pt,yscale=-0.6,xscale=0.6, every node/.style={scale=0.6}]
            
            \draw  [color={rgb, 255:red, 245; green, 166; blue, 35 }  ,draw opacity=1 ][fill={rgb, 255:red, 174; green, 210; blue, 252 }  ,fill opacity=1 ][line width=1.5]  (87,71.8) .. controls (87,56.34) and (99.54,43.8) .. (115,43.8) -- (383,43.8) .. controls (398.46,43.8) and (411,56.34) .. (411,71.8) -- (411,175.8) .. controls (411,191.26) and (398.46,203.8) .. (383,203.8) -- (115,203.8) .. controls (99.54,203.8) and (87,191.26) .. (87,175.8) -- cycle ;
            \draw   (296,130.3) .. controls (296,125.05) and (300.03,120.8) .. (305,120.8) .. controls (309.97,120.8) and (314,125.05) .. (314,130.3) .. controls (314,135.55) and (309.97,139.8) .. (305,139.8) .. controls (300.03,139.8) and (296,135.55) .. (296,130.3) -- cycle ; \draw   (298.64,123.58) -- (311.36,137.02) ; \draw   (311.36,123.58) -- (298.64,137.02) ;
            \draw    (258,130.8) -- (294,130.33) ;
            \draw [shift={(296,130.3)}, rotate = 539.25] [color={rgb, 255:red, 0; green, 0; blue, 0 }  ][line width=0.75]    (10.93,-3.29) .. controls (6.95,-1.4) and (3.31,-0.3) .. (0,0) .. controls (3.31,0.3) and (6.95,1.4) .. (10.93,3.29)   ;
            \draw    (305,172.8) -- (305,141.8) ;
            \draw [shift={(305,139.8)}, rotate = 450] [color={rgb, 255:red, 0; green, 0; blue, 0 }  ][line width=0.75]    (10.93,-3.29) .. controls (6.95,-1.4) and (3.31,-0.3) .. (0,0) .. controls (3.31,0.3) and (6.95,1.4) .. (10.93,3.29)   ;
            \draw    (315,130.8) -- (350,130.8) ;
            \draw [shift={(352,130.8)}, rotate = 180] [color={rgb, 255:red, 0; green, 0; blue, 0 }  ][line width=0.75]    (10.93,-3.29) .. controls (6.95,-1.4) and (3.31,-0.3) .. (0,0) .. controls (3.31,0.3) and (6.95,1.4) .. (10.93,3.29)   ;
            \draw   (352,131.8) .. controls (352,126.28) and (356.25,121.8) .. (361.5,121.8) .. controls (366.75,121.8) and (371,126.28) .. (371,131.8) .. controls (371,137.32) and (366.75,141.8) .. (361.5,141.8) .. controls (356.25,141.8) and (352,137.32) .. (352,131.8) -- cycle ; \draw   (352,131.8) -- (371,131.8) ; \draw   (361.5,121.8) -- (361.5,141.8) ;
            \draw    (107,60.8) -- (107,131) ;
            \draw    (107,60.8) -- (361,61.8) ;
            \draw    (361,61.8) -- (361.48,119.8) ;
            \draw [shift={(361.5,121.8)}, rotate = 269.52] [color={rgb, 255:red, 0; green, 0; blue, 0 }  ][line width=0.75]    (10.93,-3.29) .. controls (6.95,-1.4) and (3.31,-0.3) .. (0,0) .. controls (3.31,0.3) and (6.95,1.4) .. (10.93,3.29)   ;
            \draw    (361.5,141.8) -- (363,246.8) ;
            \draw    (73,245.8) -- (363,246.8) ;
            \draw    (73,245.8) -- (72.02,133.8) ;
            \draw [shift={(72,131.8)}, rotate = 449.5] [color={rgb, 255:red, 0; green, 0; blue, 0 }  ][line width=0.75]    (10.93,-3.29) .. controls (6.95,-1.4) and (3.31,-0.3) .. (0,0) .. controls (3.31,0.3) and (6.95,1.4) .. (10.93,3.29)   ;
            \draw  [dash pattern={on 0.84pt off 2.51pt}]  (206,74.8) -- (206,131.8) ;
            \draw  [dash pattern={on 0.84pt off 2.51pt}]  (206,74.8) -- (273,74.8) ;
            \draw  [dash pattern={on 0.84pt off 2.51pt}]  (273,74.8) -- (273.96,128.8) ;
            \draw [shift={(274,130.8)}, rotate = 268.98] [color={rgb, 255:red, 0; green, 0; blue, 0 }  ][line width=0.75]    (10.93,-3.29) .. controls (6.95,-1.4) and (3.31,-0.3) .. (0,0) .. controls (3.31,0.3) and (6.95,1.4) .. (10.93,3.29)   ;
            \draw    (371,131.8) -- (448,131.8) ;
            \draw [shift={(450,131.8)}, rotate = 180] [color={rgb, 255:red, 0; green, 0; blue, 0 }  ][line width=0.75]    (10.93,-3.29) .. controls (6.95,-1.4) and (3.31,-0.3) .. (0,0) .. controls (3.31,0.3) and (6.95,1.4) .. (10.93,3.29)   ;
            
            \draw    (2.9,114.3) -- (51.9,114.3) -- (51.9,146.3) -- (2.9,146.3) -- cycle  ;
            \draw (27.4,130.3) node  [color={rgb, 255:red, 0; green, 0; blue, 0 }  ,opacity=1 ] [align=left] {\begin{minipage}[lt]{30.463991699218752pt}\setlength\topsep{0pt}
            \begin{center}
            \LARGE{$z_k$}
            \end{center}
            
            \end{minipage}};
            \draw    (138.4,75.5) -- (192.4,75.5) -- (192.4,186.5) -- (138.4,186.5) -- cycle  ;
            \draw (165.4,128.5) node  [rotate=-270] [align=left] {\begin{minipage}[lt]{72.62399169921875pt}\setlength\topsep{0pt}
            \large{Dense Layer}
            \begin{center}
            \large{$\nll \rightarrow \nll$}
            \end{center}
            
            \end{minipage}};
            \draw    (227.02,79.89) -- (257.08,79.89) -- (257.08,183.91) -- (227.02,183.91) -- cycle  ;
            \draw (242.05,131.9) node  [rotate=-269.97] [align=left] {\begin{minipage}[lt]{68pt}\setlength\topsep{0pt}
            \begin{center}
            \large{LeakyReLU}
            \end{center}
            
            \end{minipage}};
            \draw (297,177) node [anchor=north west][inner sep=0.75pt]   [align=left] {\LARGE{$\alpha_i$}};
            \draw (202,179) node [anchor=north west][inner sep=0.75pt]  [rotate=90] [align=left] {\large{$i<C$}};
            \draw (280.5,81.8) node [anchor=north west][inner sep=0.75pt]  [rotate=-359] [align=left] {\large{$i=C$}};
            \draw  [color={rgb, 255:red, 245; green, 166; blue, 35 }  ,draw opacity=1 ][fill={rgb, 255:red, 174; green, 210; blue, 252 }  ,fill opacity=1 ][line width=1.5]   (199,218) -- (246,218) -- (246,243) -- (199,243) -- cycle  ;
            \draw (205,222) node [anchor=north west][inner sep=0.75pt]   [align=left] {{\Large{$C \hspace{2mm} \times$}}};
            \draw    (451,114.9) -- (500,114.9) -- (500,145.9) -- (451,145.9) -- cycle  ;
            \draw (476,130.4) node   [align=left] {\begin{minipage}[lt]{30.6pt}\setlength\topsep{0pt}
            \LARGE{$z_{k+1}$}
            \end{minipage}};
            \draw    (51.9,130.42) -- (136.4,130.85) ;
            \draw [shift={(138.4,130.86)}, rotate = 180.29] [color={rgb, 255:red, 0; green, 0; blue, 0 }  ][line width=0.75]    (10.93,-3.29) .. controls (6.95,-1.4) and (3.31,-0.3) .. (0,0) .. controls (3.31,0.3) and (6.95,1.4) .. (10.93,3.29)   ;
            \draw    (192.4,131.32) -- (225.02,131.7) ;
            \draw [shift={(227.02,131.72)}, rotate = 180.67] [color={rgb, 255:red, 0; green, 0; blue, 0 }  ][line width=0.75]    (10.93,-3.29) .. controls (6.95,-1.4) and (3.31,-0.3) .. (0,0) .. controls (3.31,0.3) and (6.95,1.4) .. (10.93,3.29)   ;
            
            \end{tikzpicture}
        }
    \caption{Surrogate network: it is based on skipconnections \cite{he2015deep}, and more precisely on an updated version of them \cite{bachlechner2020rezero}. The network consists of fully connected layers of dimension $\nll$.}
    \label{fig:surrogate}
\end{figure}
         
        In the surrogate's architecture, we found decisive the use of so-called \textit{skipconnections} \cite{he2015deep} which are now a common and good practice. It consists in adding the result of layer $\boldsymbol{i}$ to the one of layer $\boldsymbol{i-1}$ in the form $\boldsymbol{z} = \boldsymbol{z} + \textit{layer}_i \left(\boldsymbol{z}\right)$. This way we predict the increment needed to reach $\boldsymbol{z}_{k+1}$ from $\boldsymbol{z}_k$ rather than the raw output directly. We even go a step further as \cite{bachlechner2020rezero} proposed an updated version of \textit{skipconnections} that performs better: $\boldsymbol{z} = \boldsymbol{z} + \alpha_i \textit{layer}_i \left(\boldsymbol{z}\right)$ where $\alpha_i$ are trainable parameters. Our surrogate learned with and without the modulation parameters $\alpha_i$: they are clearly beneficial as we achieve better results while including them. \Cref{fig:surrogate} exposes the architecture of the surrogate network with the fully connected layers of dimension $\nll$ and the associated \textit{LeakyReLU} activations.\\

    \subsection{Learning strategy: training both networks together}
    
        As often the case when optimizing functions of several variables, performing sequential optimization by group of variables may be appealing since it reduces the search space of each optimization step. However it generally leads to a sub-optimal solution. In our case too, numerical experiments, not reported here, showed that training both the AE and the surrogate together gives better results than training the AE first and then the surrogate. Since AE's quality influences surrogate's performances, a combined training allows them to ``communicate" and ``share" information in order to more properly learn: the latent space is designed to fit the surrogate and vice versa. To do so, we define a custom loss function with a weighting parameter $\rho$ that balances between \cref{eq:AEloss} and \cref{eq:iterloss}:
        \begin{equation}
            \boldsymbol{\mathcal{L}} \left(\boldsymbol{x}_{k:k+C}\right) = \boldsymbol{\mathcal{L}}_{\textbf{AE}}\left(\boldsymbol{x}_{k+1:k+C}\right) + \rho \times \boldsymbol{\mathcal{L}}_{\textbf{Sur}}\left(\boldsymbol{x}_{k:k+C}\right)
            \label{eq:loss}
        \end{equation}
        where:
        \begin{equation*}
            \boldsymbol{\mathcal{L}}_{\textbf{AE}}\left(\boldsymbol{x}_{k+1:k+C}\right) = \frac{1}{C} \sum_{c=1}^{C} \text{MSE}\left(\mathcal{D}\left(\mathcal{E}\left(\boldsymbol{x}_{k+c}\right), \boldsymbol{x}_{k+c}\right)\right)
        \end{equation*}
        and:
        \begin{equation*}
            \boldsymbol{\mathcal{L}}_{\textbf{Sur}}\left(\boldsymbol{x}_{k:k+C}\right) = \frac{1}{C} \sum_{c=1}^{C} \text{MSE}\left(\mathcal{T}^c\left(\boldsymbol{x}_k\right), \boldsymbol{x}_{k+c}\right)
        \end{equation*}
        where $\boldsymbol{x}_{k:k+C}$ denotes the sequence $\left[\boldsymbol{x}_k,\boldsymbol{x}_{k+1}, \dots, \boldsymbol{x}_{k+C}\right]$.

\section{Data Assimilation within a latent space}
\label{sec:DA}

    Before presenting in details our ETKF-Q-Latent algorithm, we first remind basic facts of sequential DA. The upcoming sections \cref{sec:seqDA} and \cref{sec:ETKF} are strongly inspired by \cite{BocquetLecture, BocquetSiamTot, fillion}.

    \subsection{Sequential Data Assimilation}
    \label{sec:seqDA}
    
        Sequential or statistical DA is based on estimation theory and refers to a DA method for which observations are sequentially assimilated as they become available.\\
        
        Sequential data assimilation deals with the following stochastic-dynamical system:
        \begin{equation}
            \label{eq:stochasticSystem}
            \begin{cases}
                \boldsymbol{y}_k & = \mathcal{H}_k \left(\boldsymbol{x}_k\right) + \boldsymbol{\varepsilon}_k\\
                \boldsymbol{x}_k & = \mathcal{M}_k \left(\boldsymbol{x}_{k-1}\right) + \boldsymbol{\eta}_k
            \end{cases}
        \end{equation}
        where $\boldsymbol{x}_k \in \mathbb{R}^n$, $\boldsymbol{y}_k \in \mathbb{R}^p$ $\forall k \in [0,K]$, $\mathcal{M}_k$ is the nonlinear dynamical model used for time propagation of the state from time $t_{k-1}$ to time $t_k$ with the additive model error $\boldsymbol{\eta}_k$, and $\mathcal{H}_k$ is the observation operator, mapping the state from the model space to the observation space with the observation error $\boldsymbol{\varepsilon}_k$. The errors are assumed to be all unbiased, uncorrelated in time and independent from $\boldsymbol x_0$.
        
        In sequential DA, the state estimation by using this stochastic-dynamical system is obtained based on the Bayesian approach which takes into account probability distributions of the errors. Available observation is used to update the conditional probability density function (pdf) (analysis step), and then this pdf is propagated to the next time step (forecast step).
        
        A common choice for pdf is the Gaussian distribution since many processes are well described with it and it is algebraically convenient. Let us assume that in (\cref{eq:stochasticSystem}), the observation error follows a Gaussian distribution with zero mean and a covariance matrix $\boldsymbol{R}_k$ and similarly model error follows a Gaussian distribution with zero mean and a covariance matrix $\boldsymbol{Q}_k$.
        
        Using Gaussian error pdfs and under the assumption that the model and observation operators are linear (denoted by $M_k$ and $H_k$, respectively), the Kalman filter recursively finds the analysis as the conditional mean of the posteriori pdf:
        \begin{equation}
            \label{analysis_equation} 
            \boldsymbol{x}^a_k = \boldsymbol{x}^f_k + \boldsymbol{K}_k\left(\boldsymbol{y}_k-H_k\boldsymbol{x}^f_k\right).
        \end{equation}
        Here, the state $\boldsymbol{x}^f_k$ at time $t_k$ represents the model prediction from the analysis at time $t_{k-1}$, \textit{i.e.} $\boldsymbol{x}_k^f = {M}_k \left(\boldsymbol{x}^a_{k-1}\right)$and $\boldsymbol{K}_k$ denotes the Kalman gain matrix at time $k$:
        \begin{equation}
             \label{Kalman_gain_matrix}
             \boldsymbol{K}_k = \boldsymbol{P}^f_{k}H_k^T\left(\boldsymbol{R}_k + H_k \boldsymbol{P}_k^f H_k^T\right)^{-1} 
        \end{equation}
        where $\boldsymbol{P}_k^f$ is the error covariance matrix of the forecast $\boldsymbol{x}_k^f$.
        Note that the estimation given by (\cref{analysis_equation}) is also known as the \textbf{BLUE} (Best Linear Unbiased Estimator) estimate, which gives the minimum variance analysis with the choice of $ \boldsymbol{K}_k$ provided by (\cref{Kalman_gain_matrix}).
        
        Once the analysis is derived, the estimate and its error covariance matrix are propagated through time:
        \begin{align}
        \boldsymbol{x}^f_{k+1} & = M_{k+1}\left(\boldsymbol{x}^a_k\right) \\
        \boldsymbol{P}^f_{k+1} & = M_{k+1} \boldsymbol{P}^a_k M_{k+1}^T + \boldsymbol{Q}_{k+1}
        \label{eq:PropCov}
        \end{align}
        where $\boldsymbol{P}^a_k$ is the error covariance matrix of the analysis. $\boldsymbol{P}^a_k$ is derived as follows:
        \begin{equation*}
        \boldsymbol{P}^a_k = (\boldsymbol{I}_k - \boldsymbol{K}_{k} H_k) \boldsymbol{P}^f_{k}
        \end{equation*}
        with $\boldsymbol{I}_k$ being the identity matrix of order $k$.

    \subsection{Ensemble Transform Kalman Filter with additive model error: ETKF-Q}
    \label{sec:ETKF}
        Ensemble DA algorithms address downsides of the Kalman filter such as handling nonlinear models and storing and computing large matrices. For instance with a Kalman filter one has to store and manipulate error covariance matrices lying in $\mathbb{R}^{n \times n}$ which is often intractable in practice. Also, applying on both sides the model $M_k$ in \cref{eq:PropCov} to compute the forecast covariance matrix $\boldsymbol{P}^f_{k}$ is prohibitively costly. Thus, ensembles enable to approximate the forecast covariance matrix thanks to a reduced set of sample vectors. In this section, we expose a tailored version of the widely used ensemble algorithm ETKF, namely ETKF-Q \cite{fillion}. What we call ETKF-Q method precisely denotes the IEnKS-Q algorithm of \cite[Algorithm 4.1]{fillion} with parameters (L=0, K=0, S=1, G=0, one Gauss Newton loop, transform version).\\
        
        Let us consider ensemble $\boldsymbol{E}_k$ at time $k$ such that: $\boldsymbol{E}_k = \{\boldsymbol{x}_k^1, \boldsymbol{x}_k^2, \dots, \boldsymbol{x}_k^m\} \in \mathbb{R}^{n \times m}$ where $m$ is the number of members. Thus, we can empirically approximate its forecast covariance matrix:
        \begin{equation}
            \boldsymbol{P}^f_k = \frac{1}{m-1} \sum_{i=1}^m \left(\boldsymbol{x}_k^{i} - \boldsymbol{\bar{x}}_k\right)\left(\boldsymbol{x}_k^{i} - \boldsymbol{\bar{x}}_k\right)^T = \boldsymbol{X}^f_k \left(\boldsymbol{X}^f_k\right)^T
            \label{forecast_error_cov_app}
        \end{equation}
        where superscript $^i$ denotes the $i$-th member of $\boldsymbol{E}_k$ and $\boldsymbol{\bar{x}}_k$ the mean at time $k$ (\textit{i.e.} $\boldsymbol{\bar{x}}_k = \frac{1}{m} \sum_{i=1}^m \boldsymbol{x}_k^{i}$). As for $\boldsymbol{X}_k^f \in \mathbb{R}^{n \times m}$, it denotes the anomaly matrix such that $\left[\boldsymbol{X}_k^f\right]^{i} = \frac{\boldsymbol{x}_k^{i} - \boldsymbol{\bar{x}}_k}{\sqrt{m-1}}$.\\
        
        Then the analysis $\boldsymbol{x}_k^a$ can be written as:
        \begin{equation}
            \boldsymbol{x}_k^a = \boldsymbol{\bar{x}}_k + \boldsymbol{X}_k^f \boldsymbol{w}_k^a.
            \label{eq:RegularAnalysedEstimate}
        \end{equation}
        Substituting this equation into~\cref{analysis_equation} and using Sherman-Morrison-Woodbury formula (see \cite{BocquetSiam} for more details) yields: 
        \begin{equation}
        \label{ensemble_estimate}
            \boldsymbol{w}_k^a = \left(\boldsymbol{I}_{m} +  \boldsymbol{Y}_{k}
        \boldsymbol{R}^{-1} \boldsymbol{Y}_{k}\right)^{-1}
        \boldsymbol{Y}_{k}^T
        \boldsymbol{R}_k^{-1} \boldsymbol{d}_k
        \end{equation}
        being an $m$-dimensional vector with $\boldsymbol{d}_k = \boldsymbol{y}_k - \overline{\mathcal{H}\left(\boldsymbol{x}_k\right)}$. In~\cref{ensemble_estimate}, $\boldsymbol{Y}_{k}$ represents observation anomalies, i.e.
        $$
        \left[\boldsymbol{Y}_k\right]^{i} = \frac{\mathcal{H}\left(\boldsymbol{x}_k^{i}\right) - \boldsymbol{\bar{y}}_k}{\sqrt{m-1}} 
        $$
        with 
        $\boldsymbol{\bar{y}}_k = 1/m \sum_{i=1}^{m}{\mathcal{H}(\boldsymbol{x}_k^{i})}$.\\
        
        Note that the decomposition of $\boldsymbol{x}_k^a$ given by~\cref{eq:RegularAnalysedEstimate} is not unique due to rank deficient matrix $\boldsymbol{X}_k^f$. This yields an ill-defined change of variables in the ensemble space that has to be fixed with the so called gauge-fixing term (\cite{Bocquet2014}). As an alternative Fillion et al. \cite{fillion} introduce deviation matrices to overcome this problem.
        \begin{definition}
            \underline{Deviation matrix:} a deviation matrix $\boldsymbol{\Delta}$ of a symmetric semi-definite positive matrix $\boldsymbol{\Sigma}$ is an \textbf{injective} factor verifying: $\boldsymbol{\Delta} \boldsymbol{\Delta}^T = \boldsymbol{\Sigma}$.
            A deviation matrix of an ensemble is a deviation matrix of its sample covariance matrix.
        \end{definition}
        Therefore, we aim to find a deviation matrix $\boldsymbol{\Delta}_k$ of $\boldsymbol{P}_k^f$ so that formulation in \cref{eq:RegularAnalysedEstimate} yields a unique estimate $\boldsymbol{x}_k^a$. Hence, we apply \cite[Proposition 3.2]{fillion}) to $\boldsymbol{x}_k^a$ in order to ensure such a requirement. Since then, it exists a unique vector $\boldsymbol{w}_k^a \in \mathbb{R}^{m-1}$ such that:
        \begin{equation*}
            \boldsymbol{x}_k^a = \boldsymbol{\bar{x}}_k + \boldsymbol{\Delta} \boldsymbol{w}_k^a.
        \end{equation*}
        and 
        \begin{align*}
            \mathbb{E}\left[\boldsymbol{w}_k^a\right] & = \boldsymbol{0}_{m-1},\\
            \mathbb{C}\left[\boldsymbol{w}_k^a\right] & = \boldsymbol{I}_{m-1},
        \end{align*}
        where $\mathbb{E}$ and $\mathbb{C}$ are the expectation and covariance operator, respectively.\\
       
        Remains the question of calculating a deviation matrix of $\boldsymbol{P}_k^f$. Again, we rely on another Fillion et al.'s proposition (\cite[Proposition 3.3]{fillion}):
        
        \begin{proposition}
            \underline{(Deviation matrix and ensemble construction)}: Let $n,m,l \in \mathbb{N}$ such that $n \geq m, l=m-1$. Let $U_m \in \mathbb{R}^{m \times l}$ such that $\left[\frac{\boldsymbol{1}_m}{\sqrt{m}} \hspace{2mm} U_m \right] \in \mathbb{R}^{m \times m}$ be an orthonormal matrix. If $\boldsymbol{E} \in \mathbb{R}^{n \times m}$ is a full column rank ensemble, then the mean $\boldsymbol{\mu} \in \mathbb{R}^n$ and a deviation matrix $\boldsymbol{\Delta} \in \mathbb{R}^{n \times l}$ of $\boldsymbol{E}$: 
            \begin{equation}
                \left[\boldsymbol{\mu} \hspace{2mm} \boldsymbol{\Delta} \right] = \boldsymbol{E} \times \left[\frac{\boldsymbol{1}_m}{m} \hspace{2mm} \frac{U_m}{\sqrt{l}} \right].
                \label{eq:deviationMatrix}
            \end{equation}
            Conversely, if $\boldsymbol{\mu} \in \mathbb{R}^{n}$ and $\boldsymbol{\Delta} \in \mathbb{R}^{n \times l}$ then the ensemble $\boldsymbol{E} \in \mathbb{R}^{n \times m}$ defined by
            \begin{equation}
                \boldsymbol{E} = \left[\boldsymbol{\mu} \hspace{2mm} \boldsymbol{\Delta} \right] \times \left[\boldsymbol{1}_m \hspace{2mm} \sqrt{l} U_m\right]^T
                \label{eq:constructEnsemble}
            \end{equation}
            has $\boldsymbol{\mu}$ as sample mean and $\boldsymbol{\Delta} \boldsymbol{\Delta}^T$ as sample covariance matrix.
        \end{proposition}
    
        With \cref{eq:deviationMatrix}, we can compute $\boldsymbol{\Delta}_k$, a deviation matrix of $\boldsymbol{P}_k^f$ (we remind that $l=m-1$):
        \begin{equation*}
            \boldsymbol{\Delta}_k = \left[x_k^1,x_k^2,\dots,x_k^m\right] \frac{U_m}{\sqrt{m-1}}
        \end{equation*}
        Regarding $U_m \in \mathbb{R}^{m \times (m-1)}$, it is a matrix such that $\left[\frac{\boldsymbol{1}_m}{\sqrt{m}} \hspace{2mm} U_m \right]$ is orthonormal (where $\boldsymbol{1}_m$ denotes the $m$-length vector $[1,1,\dots,1]^T$). It is worth mentioning that $U_m$ can be constructed thanks to Householder's rotations.\\
        
        When propagating through time, we know that our model $\mathcal{M}_k$ is not perfect and has an intrinsic error denoted $\boldsymbol{\eta}_k$ (see \cref{eq:stochasticSystem}). However, up to now we have not included this particular knowledge in our analysis keeping the erroneous prediction as it is. Some approaches attempt to leverage this information in order to perform a model error correction and thus improve predictions' quality \cite{Sakov2018, Mitchell2015, SakovAsynch, Mandel2016, Amezcua2017}.
        
        We now come to the core of the ETKF-Q algorithm, the variant of the ETKF one which takes model error into account in the expression of the covariance matrix of $\boldsymbol{x}_k$ (here $\boldsymbol{x}_k$ denotes the real physical state):
        \begin{align*}
            \boldsymbol{x}_k & = \mathcal{M}_k\left(\boldsymbol{x}_{k-1}\right) + \boldsymbol{\eta}_k\\
            \mathbb{C}\left[\boldsymbol{x}_k|\boldsymbol{y}_{0:k-1}\right] & = \mathbb{C}\left[\mathcal{M}_k\left(\boldsymbol{x}_{k-1}\right)+\boldsymbol{\eta}_k|\boldsymbol{y}_{0:k-1}\right]
        \end{align*}
        where $\boldsymbol{y}_{0:j}$ denotes the sequence of all the observations from time $0$ to time $j$.\\
        
        We have supposed that $\boldsymbol{\eta}_k$ $\forall k \in [0,K]$ and $\boldsymbol{x}_0$ are mutually independent. Then, as $\mathcal{M}_k\left(\boldsymbol{x}_{k-1}\right)$ is a function of $\boldsymbol{x}_0$ and of $\boldsymbol{\eta}_0, \boldsymbol{\eta}_1, \dots, \boldsymbol{\eta}_{k-1}$, it comes that $\mathcal{M}_k\left(\boldsymbol{x}_{k-1}\right)$ and $\boldsymbol{\eta}_k$ are independent which yields that
        \begin{align*}
            \mathbb{C}\left[\boldsymbol{x}_k|\boldsymbol{y}_{0:k-1}\right] & = \mathbb{C}\left[\mathcal{M}_k\left(\boldsymbol{x}_{k-1}\right)|\boldsymbol{y}_{0:k-1}\right] + \mathbb{C}\left[\boldsymbol{\eta}_k|\boldsymbol{y}_{0:k-1}\right]\\
        \end{align*}
        We have also assumed that propagation and observation errors $\boldsymbol{\eta}_k$ and $\boldsymbol{\varepsilon}_k$ are mutually independent $\forall k \in [0,K]$. Then, we have $\mathbb{C}\left[\boldsymbol{\eta}_k|\boldsymbol{y}_{0:k-1}\right] = \mathbb{C}\left[\boldsymbol{\eta}_k\right] = \boldsymbol{Q}_k$.\\
        We get:
        \begin{align*}
            \mathbb{C}\left[\boldsymbol{x}_k|\boldsymbol{y}_{0:k-1}\right] & = \mathbb{C}\left[\mathcal{M}_k\left(\boldsymbol{x}_{k-1}\right)|\boldsymbol{y}_{0:k-1}\right] + \boldsymbol{Q}_k\\
        \end{align*}
        But, $\mathbb{C}\left[\mathcal{M}_k\left(\boldsymbol{x}_{k-1}\right)|\boldsymbol{y}_{0:k-1}\right]$ has been empirically approximated by $\boldsymbol{P}_k^f = \boldsymbol{\Delta}_k \boldsymbol{\Delta}_k^T$.\\
        Hence we obtain that 
        \begin{align*}
            \mathbb{C}\left[\boldsymbol{x}_k|\boldsymbol{y}_{0:k-1}\right] \approx \boldsymbol{\Delta}_k \boldsymbol{\Delta}_k^T + \boldsymbol{Q}_k.
        \end{align*}
        Deviation matrices of $\boldsymbol{\Delta}_k \boldsymbol{\Delta}_k^T + \boldsymbol{Q}_k$ are supposed to lie in $\mathbb R^{n \times n}$, but since a $n \times l$ deviation matrix is required for the next cycle, a reduction has to be performed.
        As $\boldsymbol{\Delta}_k \boldsymbol{\Delta}_k^T + \boldsymbol{Q}_k$ is symmetric (as a sum of symmetric matrices), its eigendecomposition by using the first $\ell$ $( = m-1)$ dominant eigenvectors yields $\boldsymbol{V}_k \in \mathbb{R}^{n \times (m-1)}$ and $\boldsymbol{\Lambda}_k \in \mathbb{R}^{(m-1) \times (m-1)}$ such that:
        \begin{equation*}
            \left(\boldsymbol{\Delta}_k \boldsymbol{\Delta}_k^T + \boldsymbol{Q}\right) \boldsymbol{V}_k \approx \boldsymbol{V}_k \boldsymbol{\Lambda}_k
        \end{equation*}
        One could notice that this approximation is the best one in matrix Frobenius norm.\\
        
        Therefore, a square root approximation of $\boldsymbol{\Delta}_k \boldsymbol{\Delta}_k^T + \boldsymbol{Q}_k$ is given by $\boldsymbol{V}_k \boldsymbol{\Lambda}_k^{1/2}$. We hence update $\boldsymbol{\Delta}_k = \boldsymbol{V}_k \boldsymbol{\Lambda}_k^{1/2}$.
        Then, \cref{eq:constructEnsemble} enables to update ensemble $\boldsymbol{E}_k$ according to this new statistic:
        \begin{equation*}
            \boldsymbol{E}_{k} = \bar{\boldsymbol{x}}_{k} + \boldsymbol{\Delta}_k \sqrt{m-1} U_m^T
        \end{equation*}
        Similarly, we apply \cref{eq:deviationMatrix} to the observation ensemble to produce $\boldsymbol{Y}_k$ which is analogous to the observation anomalies in the regular ETKF algorithm:
        \begin{equation*}
            \boldsymbol{Y}_k = \left[\mathcal{H}\left(\boldsymbol{x}_{k}^1\right),\mathcal{H}\left(\boldsymbol{x}_{k}^2\right),\dots,\mathcal{H}\left(\boldsymbol{x}_{k}^m\right)\right] \frac{U_m}{\sqrt{m-1}}
        \end{equation*}
        From now on, it is a straightforward application of the regular ETKF algorithm (see \cite[Section 5.3]{BocquetLecture}). In \cref{algo:ETKF-Q}, we detail the ETKF-Q algorithm with the additional assumptions that $\boldsymbol{R}_k = \boldsymbol{R}$, $\boldsymbol{Q}_k = \boldsymbol{Q}$, $\forall k \in \left[0,K\right]$. In this algorithm, we mention that operator $\mathcal{H}$ is a column-wise operator when applied to an ensemble, \textit{i.e.} $\mathcal{H}\left(\boldsymbol{E}_k\right) = \left[\mathcal{H}\left(\boldsymbol{x}_k^1\right), \mathcal{H}\left(\boldsymbol{x}_k^2\right), \dots, \left(\boldsymbol{x}_k^m\right)\right]$.
        \begin{algorithm}[t]
            \SetAlgoLined
            \caption{ETKF-Q}
            \label{algo:ETKF-Q}
                \small{
                \nonl \textbf{Inputs}: \\
                
                \nonl \qquad Observation vector $\boldsymbol{y}_0 \in \mathbb{R}^p$ \;
                
                \nonl \qquad Observation operator $\mathcal{H}: \mathbb{R}^n \rightarrow \mathbb{R}^p$ \;
                
                \nonl \qquad Obs. error covariance matrix $\boldsymbol{R} \in \mathbb{R}^{p \times p}$ \;
                
                \nonl \qquad Ensemble $\boldsymbol{E}_{0}=\{\boldsymbol{x}_{0}^1,\boldsymbol{x}_{0}^2,\dots,\boldsymbol{x}_{0}^m\} \in \mathbb{R}^{n \times m}$ \;
                
                \nonl \qquad Model operator $\mathcal{M}: \mathbb{R}^n \rightarrow \mathbb{R}^n$ \;
                
                \nonl \qquad Model error covariance matrix $\boldsymbol{Q} \in \mathbb{R}^{n \times n}$ \; 
                
                \nonl \qquad Inflation parameter $\lambda \in \mathbb{R}$ \;
                
                \nonl\texttt{\\}
                \nonl \textbf{Initialization}: \\
                
                \nonl \qquad Construct $U_m$ matrix such that  $\left[\frac{\boldsymbol{1}_m}{\sqrt{m}} \hspace{2mm} U_m \right]$ is orthonormal \; 
                
                \nonl \qquad Define $\mathcal{U} := \left[\frac{\boldsymbol{1}_m}{m} \hspace{2mm} \frac{U_m}{\sqrt{m-1}} \right]$ \;

                \nonl\texttt{\\}
                \nonl \For{$k = 1,2, ...$}{

                \nonl \qquad \qquad \qquad \qquad \textbf{Propagation step} \\ 
                $\boldsymbol{E}_k := \mathcal{M}\left(\boldsymbol{E}^a_{k-1}\right)$ \;
                
                $\left[\bar{\boldsymbol{x}}_{k} \hspace{2mm} \boldsymbol{\Delta}_k \right] := \boldsymbol{E}_k \times \mathcal{U}$ \;
                
                Calculate eigenpairs of $\left(\boldsymbol{\Delta}_k \boldsymbol{\Delta}_k^T + \boldsymbol{Q}\right)$:
                $\left(\boldsymbol{\Delta}_k \boldsymbol{\Delta}_k^T + \boldsymbol{Q}\right) \boldsymbol{V}_k \approx \boldsymbol{V}_k \boldsymbol{\Lambda}_k$ with $\boldsymbol{V}_k \in \mathbb{R}^{n \times (m-1)}$ and $\boldsymbol{\Lambda}_k \in \mathbb{R}^{(m-1) \times (m-1)}$ \;
                
                $\boldsymbol{\Delta}_k := \boldsymbol{V}_k \boldsymbol{\Lambda}_k^{1/2}$ (Update deviation matrix with model error) \;
                
                $\boldsymbol{E}_{k} := \left[\bar{\boldsymbol{x}}_{k} \hspace{2mm} \boldsymbol{\Delta}_k\right] \times \mathcal{U}^{-1}$ (Update ensemble with new statistics) \;
                
                \nonl \qquad \qquad \qquad \qquad \textbf{Analysis step} \\
                
                $\left[\bar{\boldsymbol{y}}_k \hspace{2mm} \boldsymbol{Y}_{k}\right] :=  \mathcal{H}\left(\boldsymbol{E}_k\right)\times \mathcal{U}$ \;
                
                Let $\boldsymbol{\Omega}_k \in \mathbb{R}^{(m-1) \times (m-1)}$ such that: $ \boldsymbol{\Omega}_k \boldsymbol{\Omega}_k^T = \left(\boldsymbol{I}_{m-1} + \boldsymbol{Y}_k^T
                \boldsymbol{R}^{-1} \boldsymbol{Y}_{k}\right)^{-1}$ \;
                $\boldsymbol{w}_k^a := \boldsymbol{\Omega}_k \boldsymbol{\Omega}_k^T \boldsymbol{Y}_{k} \boldsymbol{R}^{-1} \left(\boldsymbol{y}_k - \bar{\boldsymbol{y}}_k\right)$ \;
                
                $\boldsymbol{E}^a_{k} := \bar{\boldsymbol{x}}_{k}\boldsymbol{1}_m^T + \lambda \times \boldsymbol{\Delta}_k \left(\boldsymbol{w}^a_k \boldsymbol{1}_m + \sqrt{m-1}\boldsymbol{\Omega}_k\right)$ \;
                }
          }
        \end{algorithm}

    \subsection{ETKF-Q-Latent algorithm}
    \label{sec:latentETKF-Q}
    
        Our goal with the ETKF-Q-L algorithm is to perform DA analysis within the latent space of our autoencoder. Indeed, we now assume the existence of a $\dimx$-dimensional system possessing a latent representation of lower dimension $\nll$. From now on, variable $\dimx$ refers to the full space dimension whereas notation $\nll$ denotes the latent space dimension. \Cref{algo:LatentETKF-Q} exposes the changes we made to do so. Here again operator $\mathcal{H}$ is a column-wise operator when applied to an ensemble.
        We highlight that ensemble $\boldsymbol{E}_0 \in \mathbb{R}^{\dimx \times m}$ is first encoded into ensemble $\boldsymbol{Z}_0 \in \mathbb{R}^{\nll \times m}$ and then all computations happen within the latent space. In order to calculate the misfit vector $\boldsymbol{d}_k = \boldsymbol{y}_k - \overline{\mathcal{H}\left(\boldsymbol{x}_k\right)}$, first the decoder $\mathcal{D}$ is used to map the ensemble from the latent space to the full space, then the observation operator $\mathcal{H}$ maps the ensemble from the full space to the observation space. Therefore, we do not need to perform any operation with the observations. They stay in their original space.
        Since time propagation is performed in the latent space, we no longer refer to matrix $\boldsymbol{Q}$ but we rather introduce $\boldsymbol{Q}_{\nll}$. Instead of using $\boldsymbol{\Delta}$ to represent a deviation matrix, we refer to it as $\boldsymbol{\Gamma}$ in the case of the latent algorithm. 
        For simplicity, we assume that $\boldsymbol{R}=\boldsymbol{\sigma}_R^2 \boldsymbol{I}_p$ and $\boldsymbol{Q}_{\nll} = \boldsymbol{\sigma}_{Q_{\nll}}^2 \boldsymbol{I}_{\nll}$. Since we cannot characterize the error committed by our surrogate network, the model correction step needs a tuned parameter $\boldsymbol{\sigma}_{Q_{\nll}}$ that embodies the unknown surrogate error.\\
        
        \begin{algorithm}[t]
            \SetAlgoLined
                \small{
                \nonl \textbf{Inputs}: \\
                \nonl \qquad Observation vector $\boldsymbol{y}_0 \in \mathbb{R}^p$ \;
                
                \nonl \qquad Observation operator $\mathcal{H} \circ \mathcal{D}:
                \mathbb{R}^{\nll} \rightarrow \mathbb{R}^{\dimx} \rightarrow \mathbb{R}^p$ \;
                
                \nonl \qquad Obs. error covariance matrix $\boldsymbol{R} \in \mathbb{R}^{p \times p}$ \;
                
                \nonl \qquad Ensemble $\boldsymbol{E}_{0}=\{\boldsymbol{x}_{0}^1,\boldsymbol{x}_{0}^2,\dots,\boldsymbol{x}_{0}^m\} \in \mathbb{R}^{\dimx \times m}$ \;
                
                \nonl \qquad Surrogate model $\mathcal{S}: \mathbb{R}^{\nll} \rightarrow \mathbb{R}^{\nll}$ \;
                
                \nonl \qquad Model error covariance matrix $\boldsymbol{Q}_{\nll} \in \mathbb{R}^{\nll \times \nll}$ \;
                
                \nonl \qquad Encoder $\mathcal{E}: \mathbb{R}^{\dimx} \rightarrow \mathbb{R}^{\nll}$ \;
                
                \nonl \qquad Inflation parameter $\lambda \in \mathbb{R}$ \;
                
                \nonl\texttt{\\}
                \nonl \textbf{Initialization}: \\
                \nonl \qquad Construct $U_m$ matrix such that  $\left[\frac{\boldsymbol{1}_m}{\sqrt{m}} \hspace{2mm} U_m \right]$ is orthonormal \; \nonl\qquad Define $\mathcal{U}:= \left[\frac{\boldsymbol{1}_m}{m} \hspace{2mm} \frac{U_m}{\sqrt{m-1}}\right]$ \;
               \nonl \qquad Encode ensemble $\boldsymbol{E}_0$: $\boldsymbol{Z}_0 := \mathcal{E}\left(\boldsymbol{E}_0\right) \in \mathbb{R}^{\nll \times m}$ \;
              
              \nonl\texttt{\\}
              \nonl \For{$k = 1,2, ...$}{
              \nonl \qquad \qquad \qquad \qquad \textbf{Propagation Step} \\
              
              $\boldsymbol{Z}_k := \mathcal{S}\left(\boldsymbol{Z}_{k-1}^a\right)$ \;
              
              $\left[\bar{\boldsymbol{z}}_k \hspace{2mm} \boldsymbol{\Gamma}_k\right] := \boldsymbol{Z}_k \times \mathcal{U}$ \;
              
              Calculate eigenpairs of $\left(\boldsymbol{\Gamma_k} \boldsymbol{\Gamma_k}^T + \boldsymbol{Q}_{\nll}\right)$:
              $\left(\boldsymbol{\Gamma_k} \boldsymbol{\Gamma_k}^T + \boldsymbol{Q}_{\nll}\right)\boldsymbol{V}_k \approx \boldsymbol{V}_k \boldsymbol{\Lambda}_k$ with $\boldsymbol{V}_k \in \mathbb{R}^{\nll \times (m-1)}$ and  $\boldsymbol{\Lambda}_k \in \mathbb{R}^{(m-1) \times (m-1)}$ \; 
              
              $\boldsymbol{\Gamma}_k := \boldsymbol{V}_k \boldsymbol{\Lambda}^{1/2}$ (Update deviation matrix with model error) \;
              
              $\boldsymbol{Z}_k := \left[\bar{\boldsymbol{z}}_k \hspace{2mm} \boldsymbol{\Gamma}_k\right] \times \mathcal{U}^{-1}$ (Update ensemble with new statistics) \;
              
              \nonl \qquad \qquad \qquad \qquad \textbf{Analysis step} \\
              
              $\left[\bar{\boldsymbol{y}}_k \hspace{2mm} \boldsymbol{Y}_{k}\right] := \mathcal{H}\left(\mathcal{D}\left(\boldsymbol{Z}_k\right)\right) \times \mathcal{U}$ \; 
              
              Let $\boldsymbol{\Omega}_k \in \mathbb{R}^{(m-1) \times (m-1)}$ such that: $ \boldsymbol{\Omega}_k \boldsymbol{\Omega}_k^T := \left(\boldsymbol{I}_{m-1} + \boldsymbol{Y}_k^T \boldsymbol{R}^{-1} \boldsymbol{Y}_{k}\right)^{-1}$ \;
              
              $\boldsymbol{w}_k^a := \boldsymbol{\Omega}_k \boldsymbol{\Omega}_k^T \boldsymbol{Y}_{k} \boldsymbol{R}^{-1} \left(\boldsymbol{y}_k - \bar{\boldsymbol{y}}_k\right)$ \;
              
              $\boldsymbol{Z}^a_{k} := \bar{\boldsymbol{z}}_{k}\boldsymbol{1}_m^T + \lambda \times \boldsymbol{\Gamma}_k \left(\boldsymbol{w}^a_k \boldsymbol{1}_m + \sqrt{m-1}\boldsymbol{\Omega}_k\right)$ \;
              }
              }
            \caption{ETKF-Q-Latent}
            \label{algo:LatentETKF-Q}
        \end{algorithm}
        
\begin{figure}
    \centering
        \scalebox{1.0}{
            \tikzset{every picture/.style={line width=0.75pt}} 
            \begin{tikzpicture}[x=0.75pt,y=0.75pt,yscale=-0.7,xscale=0.7, every node/.style={scale=0.7}]
        
            \draw    (34.3,131.4) -- (81.3,131.4) ;
            \draw [shift={(83.3,131.4)}, rotate = 180] [color={rgb, 255:red, 0; green, 0; blue, 0 }  ][line width=0.75]    (10.93,-3.29) .. controls (6.95,-1.4) and (3.31,-0.3) .. (0,0) .. controls (3.31,0.3) and (6.95,1.4) .. (10.93,3.29)   ;
            \draw    (124.3,131.4) -- (200.05,131.28) ;
            \draw [shift={(202.05,131.27)}, rotate = 539.9100000000001] [color={rgb, 255:red, 0; green, 0; blue, 0 }  ][line width=0.75]    (10.93,-3.29) .. controls (6.95,-1.4) and (3.31,-0.3) .. (0,0) .. controls (3.31,0.3) and (6.95,1.4) .. (10.93,3.29)   ;
            \draw    (238.3,131.4) -- (285.3,131.4) ;
            \draw [shift={(287.3,131.4)}, rotate = 180] [color={rgb, 255:red, 0; green, 0; blue, 0 }  ][line width=0.75]    (10.93,-3.29) .. controls (6.95,-1.4) and (3.31,-0.3) .. (0,0) .. controls (3.31,0.3) and (6.95,1.4) .. (10.93,3.29)   ;
            \draw    (306.3,187.4) -- (306.3,147.4) ;
            \draw [shift={(306.3,145.4)}, rotate = 450] [color={rgb, 255:red, 0; green, 0; blue, 0 }  ][line width=0.75]    (10.93,-3.29) .. controls (6.95,-1.4) and (3.31,-0.3) .. (0,0) .. controls (3.31,0.3) and (6.95,1.4) .. (10.93,3.29)   ;
            \draw    (331.3,131.4) -- (382.05,131.28) ;
            \draw [shift={(384.05,131.27)}, rotate = 539.86] [color={rgb, 255:red, 0; green, 0; blue, 0 }  ][line width=0.75]    (10.93,-3.29) .. controls (6.95,-1.4) and (3.31,-0.3) .. (0,0) .. controls (3.31,0.3) and (6.95,1.4) .. (10.93,3.29)   ;
            \draw    (403.3,89.4) -- (403.05,113.27) ;
            \draw    (103.3,89.4) -- (403.3,89.4) ;
            \draw    (103.3,89.4) -- (103.3,113.4) ;
            \draw [shift={(103.3,115.4)}, rotate = 270] [color={rgb, 255:red, 0; green, 0; blue, 0 }  ][line width=0.75]    (10.93,-3.29) .. controls (6.95,-1.4) and (3.31,-0.3) .. (0,0) .. controls (3.31,0.3) and (6.95,1.4) .. (10.93,3.29)   ;
            
            \draw (10,127) node [anchor=north west][inner sep=0.75pt]   [align=left] {\large{$\boldsymbol{x}_k$}};
            \draw (45,110) node [anchor=north west][inner sep=0.75pt]   [align=left] {{$\mathcal{E}\left(.\right)$}};
            \draw (96,127) node [anchor=north west][inner sep=0.75pt]   [align=left] {\large{$\boldsymbol{z}_k$}};
            \draw (124,110) node [anchor=north west][inner sep=0.75pt]   [align=left] {{\tiny Model Error Correction}};
            \draw (212,122) node [anchor=north west][inner sep=0.75pt]   [align=left] {\large{$\boldsymbol{z}_k^{'}$}};
            \draw (298,122) node [anchor=north west][inner sep=0.75pt]   [align=left] {\large{$\boldsymbol{z}_k^{''}$}};
            \draw (300,191) node [anchor=north west][inner sep=0.75pt]   [align=left] {\large{$\boldsymbol{y}_k$}};
            \draw (340,110) node [anchor=north west][inner sep=0.75pt]   [align=left] {{$\mathcal{S}\left(.\right)$}};
            \draw (392,127) node [anchor=north west][inner sep=0.75pt]   [align=left] {\large{$\boldsymbol{z}_{k+1}$}};
            \draw (230,69) node [anchor=north west][inner sep=0.75pt] [align=left] {Loop};
            
            \end{tikzpicture}
        }
        \caption{Outline of the Data Assimilation framework that includes the trained neural networks: $\boldsymbol{z}_k$ denotes the encoded input at time $k$, $\boldsymbol{z}_k^{'}$ represents $\boldsymbol{z}_k$ corrected with respect to the model error, $\boldsymbol{z}_k^{''}$ is the estimate yielded by assimilating observation $\boldsymbol{y}_k$, and $\mathcal{E\left(.\right)}$ and $\mathcal{S}\left(.\right)$ denote the encoder and the surrogate operator, respectively.}
\label{fig:architecture}
\end{figure}
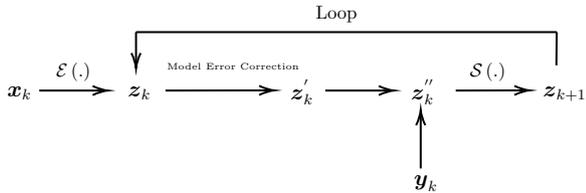

        In order to picture the overall architecture of our DA framework, we can refer to \cref{fig:architecture}.

        The strong difference between \cref{algo:ETKF-Q} and \cref{algo:LatentETKF-Q}, is the reduction of the computational space from $\mathbb{R}^{\dimx}$ to $\mathbb{R}^{\nll}$, which straightforwardly reduces both the computational cost and the memory storage. In practice, our latent space is $10$ times smaller than our full space (see \cref{sec:assessingPerf} for detailed results).
        
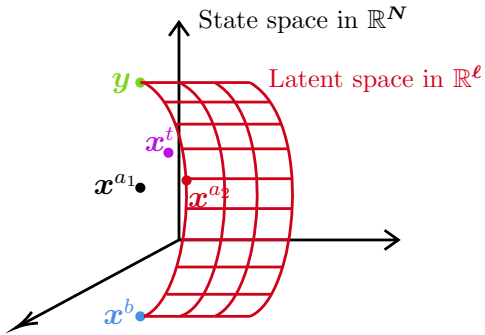
\begin{figure}
    \centering
    \scalebox{1.4}{
    \tikzset{every picture/.style={line width=0.75pt}} 
    
    \begin{tikzpicture}[x=0.75pt,y=0.75pt,yscale=-0.7,xscale=0.7, every node/.style={scale=0.7}]
    
    \draw  (199.88,190.33) -- (311.61,190.33)(199.88,78.6) -- (199.88,190.33) -- cycle (304.61,185.33) -- (311.61,190.33) -- (304.61,195.33) (194.88,85.6) -- (199.88,78.6) -- (204.88,85.6)  ;
    \draw    (199.88,190.33) -- (118.64,234.38) ;
    \draw [shift={(116.88,235.33)}, rotate = 331.53] [color={rgb, 255:red, 0; green, 0; blue, 0 }  ][line width=0.75]    (10.93,-3.29) .. controls (6.95,-1.4) and (3.31,-0.3) .. (0,0) .. controls (3.31,0.3) and (6.95,1.4) .. (10.93,3.29)   ;
    
    \draw  [color={rgb, 255:red, 126; green, 211; blue, 33 }  ,draw opacity=1 ][fill={rgb, 255:red, 126; green, 211; blue, 33 }  ,fill opacity=1 ] (178.33,109.54) .. controls (178.33,108.51) and (179.17,107.67) .. (180.19,107.67) .. controls (181.22,107.67) and (182.05,108.51) .. (182.05,109.54) .. controls (182.05,110.58) and (181.22,111.42) .. (180.19,111.42) .. controls (179.17,111.42) and (178.33,110.58) .. (178.33,109.54) -- cycle ;
    \draw  [fill={rgb, 255:red, 0; green, 0; blue, 0 }  ,fill opacity=1 ] (178.33,163.54) .. controls (178.33,162.51) and (179.17,161.67) .. (180.19,161.67) .. controls (181.22,161.67) and (182.05,162.51) .. (182.05,163.54) .. controls (182.05,164.58) and (181.22,165.42) .. (180.19,165.42) .. controls (179.17,165.42) and (178.33,164.58) .. (178.33,163.54) -- cycle ;
    \draw  [color={rgb, 255:red, 74; green, 144; blue, 226 }  ,draw opacity=1 ][fill={rgb, 255:red, 74; green, 144; blue, 226 }  ,fill opacity=1 ] (178.33,229.54) .. controls (178.33,228.51) and (179.17,227.67) .. (180.19,227.67) .. controls (181.22,227.67) and (182.05,228.51) .. (182.05,229.54) .. controls (182.05,230.58) and (181.22,231.42) .. (180.19,231.42) .. controls (179.17,231.42) and (178.33,230.58) .. (178.33,229.54) -- cycle ;
    \draw [color={rgb, 255:red, 208; green, 2; blue, 27 }  ,draw opacity=1 ]   (182.05,109.54) .. controls (215.09,128.89) and (206.42,219.56) .. (182.05,229.54) ;
    \draw [color={rgb, 255:red, 208; green, 2; blue, 27 }  ,draw opacity=1 ]   (200.05,109.54) .. controls (235.71,127.56) and (225.71,223.56) .. (200.05,229.54) ;
    \draw [color={rgb, 255:red, 208; green, 2; blue, 27 }  ,draw opacity=1 ]   (218.05,109.54) .. controls (251.09,128.89) and (242.42,219.56) .. (218.05,229.54) ;
    \draw [color={rgb, 255:red, 208; green, 2; blue, 27 }  ,draw opacity=1 ]   (236.05,109.54) .. controls (269.09,128.89) and (260.42,219.56) .. (236.05,229.54) ;
    \draw [color={rgb, 255:red, 208; green, 2; blue, 27 }  ,draw opacity=1 ]   (182.05,229.54) -- (200.05,229.54) ;
    \draw [color={rgb, 255:red, 208; green, 2; blue, 27 }  ,draw opacity=1 ]   (200.05,229.54) -- (218.05,229.54) ;
    \draw [color={rgb, 255:red, 208; green, 2; blue, 27 }  ,draw opacity=1 ]   (218.05,229.54) -- (236.05,229.54) ;
    \draw [color={rgb, 255:red, 208; green, 2; blue, 27 }  ,draw opacity=1 ]   (180.19,109.54) -- (200.05,109.54) ;
    \draw [color={rgb, 255:red, 208; green, 2; blue, 27 }  ,draw opacity=1 ]   (218.05,109.54) -- (200.05,109.54) ;
    \draw [color={rgb, 255:red, 208; green, 2; blue, 27 }  ,draw opacity=1 ]   (218.05,109.54) -- (236.05,109.54) ;
    \draw [color={rgb, 255:red, 208; green, 2; blue, 27 }  ,draw opacity=1 ]   (192.42,119.56) -- (245.76,119.56) ;
    \draw [color={rgb, 255:red, 208; green, 2; blue, 27 }  ,draw opacity=1 ]   (198.42,130.89) -- (251.76,130.89) ;
    \draw [color={rgb, 255:red, 208; green, 2; blue, 27 }  ,draw opacity=1 ]   (201.76,143.56) -- (255.76,143.56) ;
    \draw [color={rgb, 255:red, 208; green, 2; blue, 27 }  ,draw opacity=1 ]   (203.76,158.89) -- (257.09,158.89) ;
    \draw [color={rgb, 255:red, 208; green, 2; blue, 27 }  ,draw opacity=1 ]   (203.09,174.22) -- (257.76,174.22) ;
    \draw [color={rgb, 255:red, 208; green, 2; blue, 27 }  ,draw opacity=1 ]   (199.09,204.22) -- (252.42,204.22) ;
    \draw [color={rgb, 255:red, 208; green, 2; blue, 27 }  ,draw opacity=1 ]   (193.76,218.22) -- (247.09,218.22) ;
    \draw [color={rgb, 255:red, 208; green, 2; blue, 27 }  ,draw opacity=1 ]   (202.55,190.17) -- (256.22,190.17) ;
    \draw  [color={rgb, 255:red, 208; green, 2; blue, 27 }  ,draw opacity=1 ][fill={rgb, 255:red, 208; green, 2; blue, 27 }  ,fill opacity=1 ] (201.9,159.76) .. controls (201.9,158.73) and (202.73,157.89) .. (203.76,157.89) .. controls (204.78,157.89) and (205.61,158.73) .. (205.61,159.76) .. controls (205.61,160.8) and (204.78,161.64) .. (203.76,161.64) .. controls (202.73,161.64) and (201.9,160.8) .. (201.9,159.76) -- cycle ;
    \draw  [color={rgb, 255:red, 189; green, 16; blue, 224 }  ,draw opacity=1 ][fill={rgb, 255:red, 189; green, 16; blue, 224 }  ,fill opacity=1 ] (192.67,145.54) .. controls (192.67,144.51) and (193.5,143.67) .. (194.53,143.67) .. controls (195.55,143.67) and (196.38,144.51) .. (196.38,145.54) .. controls (196.38,146.58) and (195.55,147.42) .. (194.53,147.42) .. controls (193.5,147.42) and (192.67,146.58) .. (192.67,145.54) -- cycle ;
    
    \draw (208.33,69) node [anchor=north west][inner sep=0.75pt]   [align=left] {{State space in $\mathbb{R}^{\dimx}$}};
    \draw (164,103.33) node [anchor=north west][inner sep=0.75pt] [align=left] {\textcolor[rgb]{0.49,0.83,0.13}{{\large{$\boldsymbol{y}$}}}};
    \draw (155,155) node [anchor=north west][inner sep=0.75pt]  [align=left] {{\large{$\boldsymbol{x}^{a_1}$}}};
    \draw (160,219) node [anchor=north west][inner sep=0.75pt]  [align=left] {{\textcolor[rgb]{0.29,0.56,0.89}{\large{$\boldsymbol{x}^b$}}}};
    \draw (244,99.2) node [anchor=north west][inner sep=0.75pt] [align=left] {{\textcolor[rgb]{0.82,0.01,0.11}{Latent space in $\mathbb{R}^{\nll}$}}};
    
    \draw (203,161.33) node [anchor=north west][inner sep=0.75pt] [align=left] {{\textcolor[rgb]{0.82,0.01,0.11}{\large{$\boldsymbol{x}^{a_2}$}}}};
    
    \draw (181,131) node [anchor=north west][inner sep=0.75pt] [align=left] {{\textcolor[rgb]{0.74,0.06,0.88}{\large{$\boldsymbol{x}^t$}}}};
    
    \end{tikzpicture}
    }
    \caption{Comparison between full space DA and latent space DA (temporal subscript is dropped here). Variables $\boldsymbol{x}^b$, $\boldsymbol{x}^{a_1}$, $\boldsymbol{x}^{a_2}$ and $\boldsymbol{x}^t$ denote the background knowledge, the full space estimate, the latent estimate and the ground truth state, respectively. We also assume $\mathcal{H}=\boldsymbol{I}_{\dimx}$.}
    \label{fig:illustration}
\end{figure}

        Numerical experiments also show an accuracy improvement when the assimilation is performed within the latent space. Indeed, decoding the analysis lying in the latent space outputs a state that is more likely to be on the system's trajectory while the linear analysis in the full space may deviate from it. \Cref{fig:illustration} details how latent DA works compared to the regular DA: since a latent dynamics exists in $\mathbb{R}^{\nll}$, latent DA leverages the nonlinear transformation provided by the encoder whereas full space DA might not capture the intrinsic dynamics and yields a poorer estimate.

\section{Numerical experiments}
\label{section:numerical experiments}

    \subsection{Choice of a physical system: The augmented Lorenz 96 model}

        The Lorenz 96 model~\cite{Lorenz96} is widely used as a dynamical system \cite{Brajard_2020, Vlachas2018, pawar2020data} in ensemble data assimilation in particular for weather prediction.\\
        
        It is defined as follows:
        \begin{equation}
            \frac{\mathrm d\boldsymbol{x_{[i]}}}{\mathrm dt} = \left(\boldsymbol{x}_{[i+1]}-\boldsymbol{x}_{[i-2]}\right)\boldsymbol{x}_{[i-1]} - \boldsymbol{x}_{[i]} + F, \quad \forall i = 1,\dots,L
            \label{eq:lorenz96}
        \end{equation}
        with $\boldsymbol{x}_{[-1]} = \boldsymbol{x}_{[L-1]}$, $\boldsymbol{x}_{[0]} = \boldsymbol{x}_{[L]}$, $\boldsymbol{x}_{[L+1]} = \boldsymbol{x}_{[1]}$, and $L \geq 4$. Subscript $[i]$ denotes the $i$-th variable.\\
        
        In this equation, quadratic terms represent the advection that conserves the total energy, linear term represents the damping through which the energy decreases, and the constant term represents external forcing keeping the total energy away from zero. The L variables may be thought of as values of some atmospheric quantity in L sectors of a latitude circle.
        For $F=8$, the system is known to have a chaotic behavior~\cite{Lorenz96}.

        We consider the Lorenz 96 dynamics with $L=40$ and then construct an augmented model based on this latent space representation. Doing so, we guarantee the existence of a latent space in which the observed dynamical system can be accurately expressed. Hence, by construction our augmented Lorenz dynamics has a latent representation of dimension $40$. 
    
\begin{figure}[t]
            \centering
                \scalebox{0.25}{
                    \begin{tabular}{c c}
                    \includegraphics{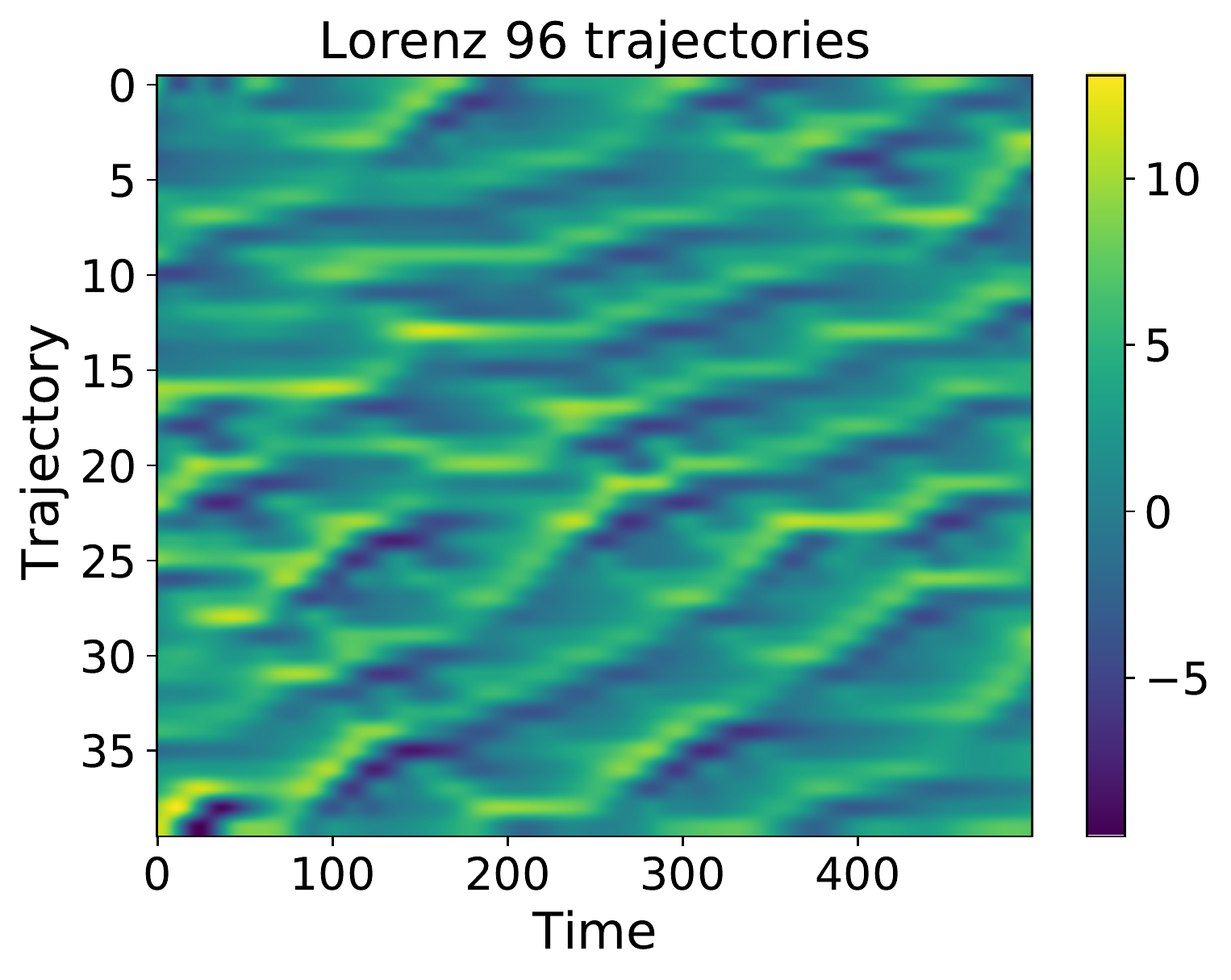} & \includegraphics{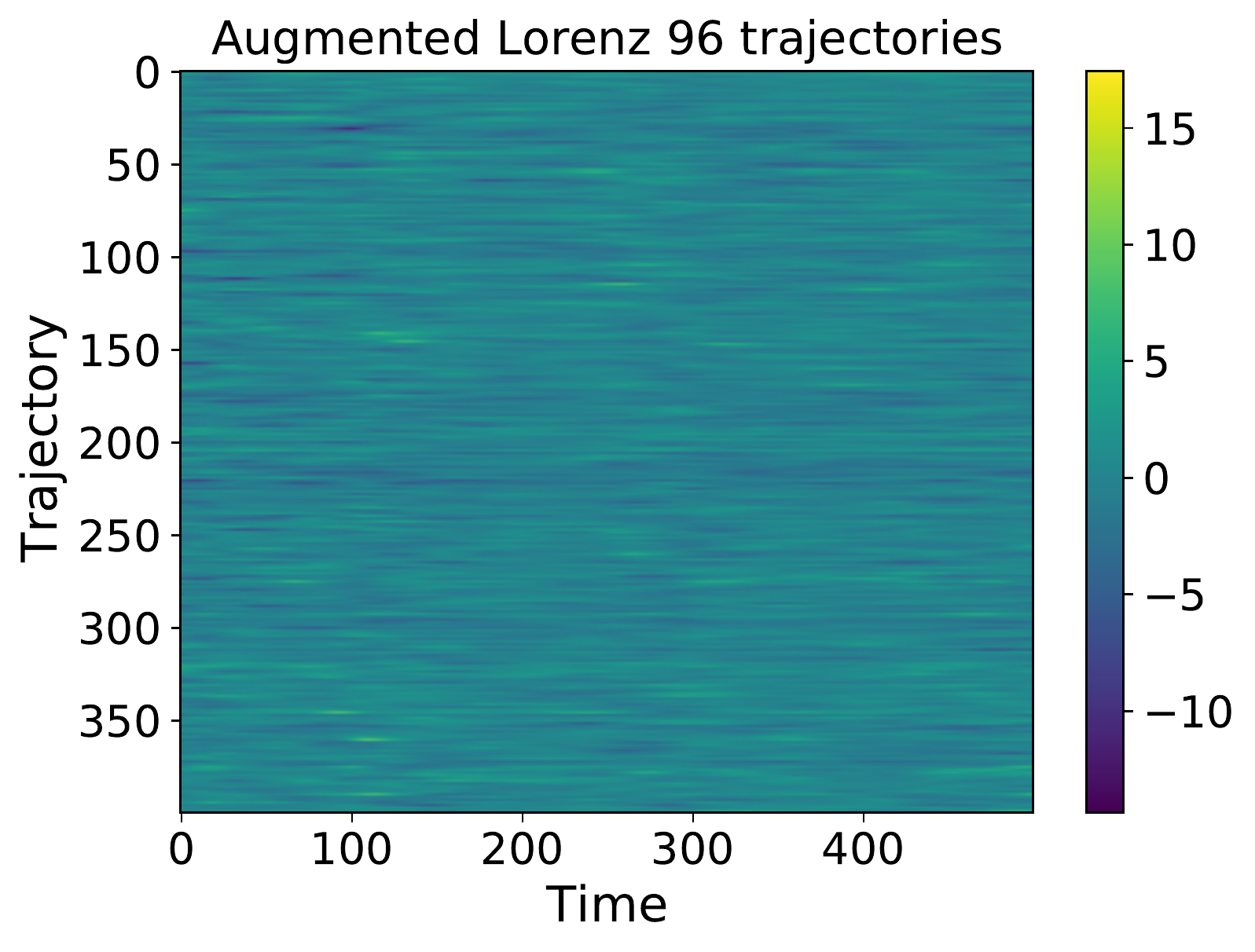}\\ 
                    \includegraphics{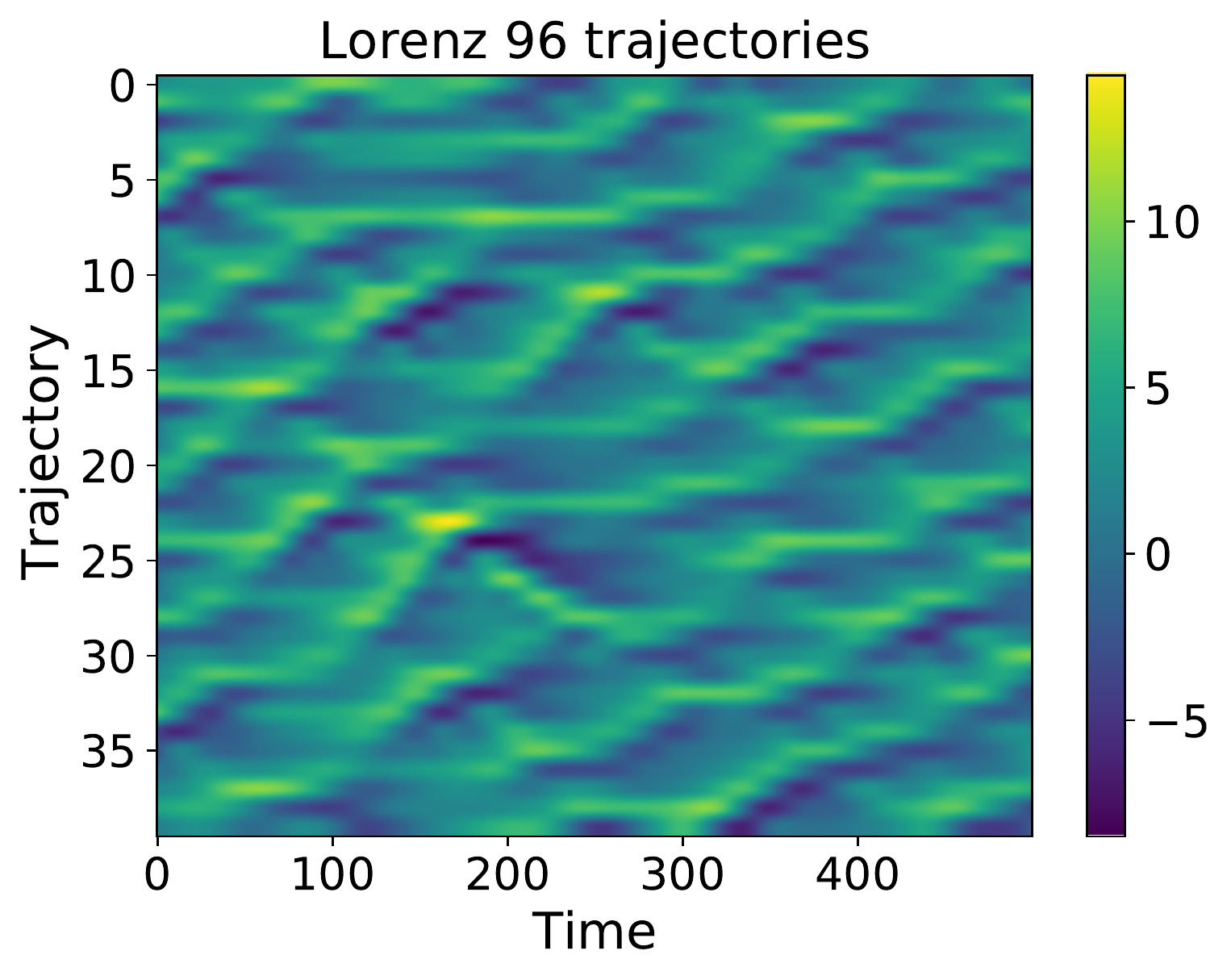} & \includegraphics{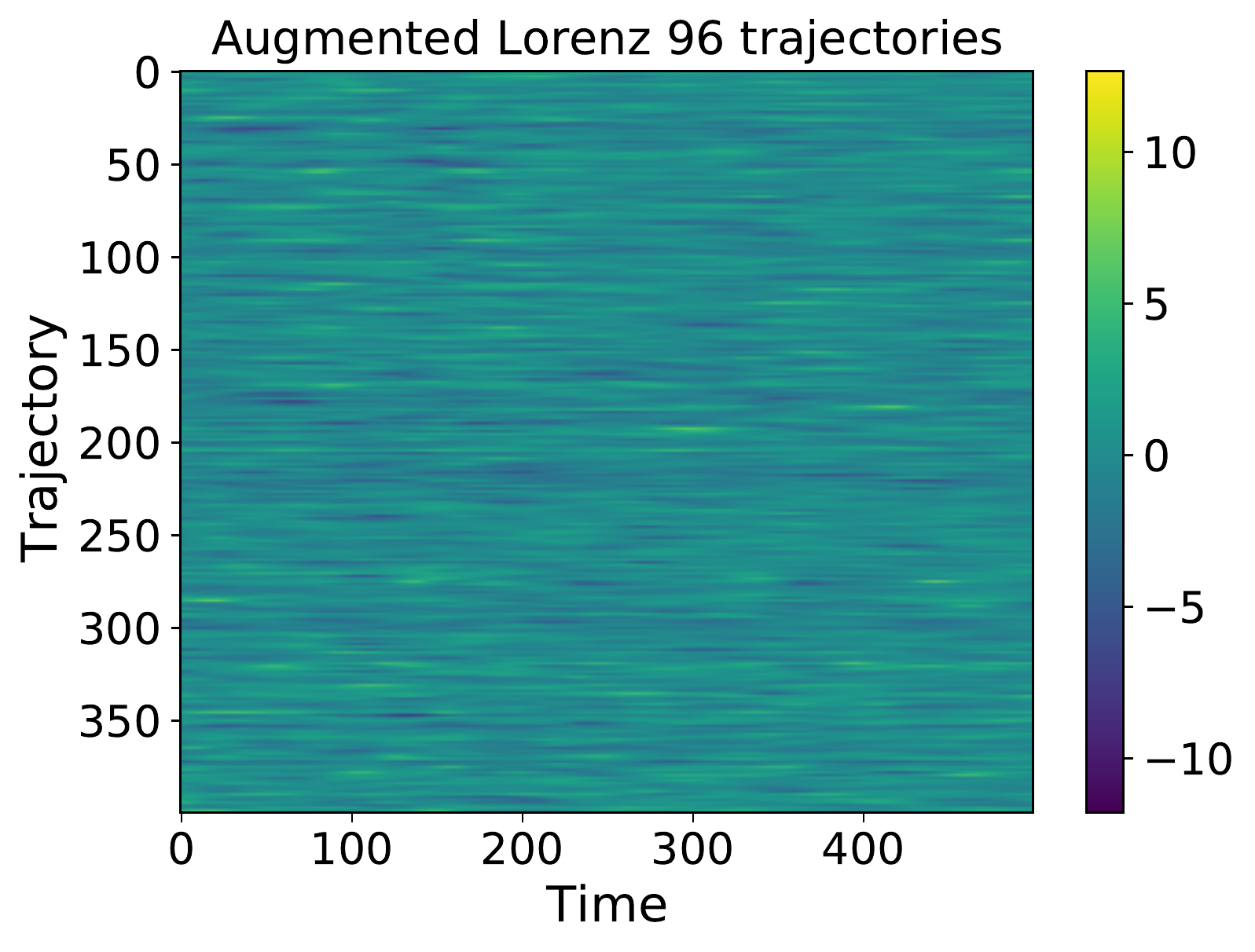}
                    \end{tabular}}
                    \caption{An illustration of two $40$-variables Lorenz dynamics (top and bottom left) and their associated augmented Lorenz models in dimension $400$ (top and bottom right). By construction, the augmented Lorenz systems has a latent representation in dimension $40$.}
                    \label{fig:AugmentedLorenz}
        \end{figure}

        The definition of the augmented Lorenz system is given as follows:
        \begin{definition}\label{def:augmentedLorenz}
            \underline{Augmented Lorenz model:} it is a $\dimx$-dimensional system for which there exists a function $\boldsymbol{\mathcal{F}}$ such that $\boldsymbol{\mathcal{F}}$ transforms the augmented Lorenz model into a $\nll$-dimensional system $\left(\nll<<\dimx\right)$ that follows Lorenz 96 equations.
        \end{definition}
        Let us consider simulations of a $\nll(=40)$-dimensional Lorenz 96 state generated by integrating the well-known Lorenz equations (with a $4^{\text{th}}$ order Runge-Kutta scheme for instance). Then, we apply a non-linear function $\boldsymbol{F}: \mathbb{R}^{\nll} \rightarrow \mathbb{R}^{\dimx}$ to generate the augmented Lorenz 96 state of dimension $\dimx$.
        In applying the nonlinear function, the $\nll$-dimensional state is first mapped
        into a ${\dimx}$-dimensional state via an orthonormal matrix $\boldsymbol{O} \in \mathbb{R}^{\nll \times \dimx}$
        and then an element-wise non-linear function $\boldsymbol{f}$ (which is basically an invertible $3^{\text{rd}}$ degree polynomial)
        is applied, i.e. 
        $$
        \boldsymbol{F}: \boldsymbol{x} \rightarrow \boldsymbol{f}\left(\boldsymbol{O}\left(\boldsymbol{x}\right)\right)
        $$
        \Cref{fig:AugmentedLorenz} shows two Lorenz 96 dynamics (generated from two very close but different initial points) in dimension $40$ and their associated augmented Lorenz 96 models in $\mathbb{R}^{400}$. This figure helps in visualising how these dynamics behave in their respective spaces. 
        
    \subsection{Networks' architectures}
    
        The encoder is made of $4$ fully connected layers each one followed by a $0.2$ slope \textit{LeakyReLU} activation except for the last layer whose activation function is \textit{tanh}. These $4$ fully connected layers maps from $\mathbb{R}^{400}$ to $\mathbb{R}^{40}$ going by dimensions $300, 200$ and $150$. The decoder performs the reverse operation but there is no activation function in the last layer. Several variants have been tested but these consecutive layers give the best achievements.\\

        The surrogate network consists of $5$ fully connected layers each one being followed by a $0.2$ slope \textit{LeakyReLU} activation except the last one. Data remain in $\mathbb{R}^{40}$ through this network.

    \subsection{Neural Networks training}
        
        We have generated $1000$ Lorenz 96 simulations with $500$ time steps and $40$ trajectories each, \textit{i.e.} a thousand of $500 \times 40$ images. We precise that in this paper a trajectory denotes the time evolution of one variable $\boldsymbol{x}_{[i]}$ (see \cref{eq:lorenz96}), \textit{i.e.} of a row in a $500 \times 40$ image, whereas a simulation represents the image itself. 
    
        All simulations come from the same distribution defined as follows (Python code):
        \vspace{-0.2cm}
        \newsavebox{\mybox}
        \begin{lrbox}{\mybox}
        \begin{lstlisting}[autogobble=True, language=python,numbers=left,stepnumber=1]
        X = np.zeros((time, nb_sim, dim))
        
        for i in range(nb_sim):
            x0 = np.random.randn(dim) * 0.01 + F0
            initial_perturbation = np.random.randn(dim)
            x0 += initial_perturbation
            X[:, i] = new_trajectory(x0, dim, burn, time, F0, deltaT)
        \end{lstlisting}
        
        \end{lrbox}
        
        \begin{center}
            \hspace*{0.2cm}\scalebox{0.65}{\usebox{\mybox}}
        \end{center}
        where \emph{x0} denotes the initial state, \emph{F0} the forcing term in Lorenz $96$ system (denoted $F$ in \cref{eq:lorenz96}), \emph{dim} is the number of variables (here $40$), \emph{burn} the number of burned states, \emph{time} the number of time steps, \emph{nb\_sim} the number of simulations (here $1000$) and \emph{deltaT} represents a single time step that is $0.01$ in our case.
        
        We have then transformed the $1000$ Lorenz $96$ simulations into $1000$ augmented Lorenz $96$ ones, \textit{i.e.} into a thousand $400$-dimensional data.\\
        
\begin{figure}[t]
            \centering
                \scalebox{1.3}{
                    \begin{tabular}{c c}
                    \includegraphics[scale=0.2]{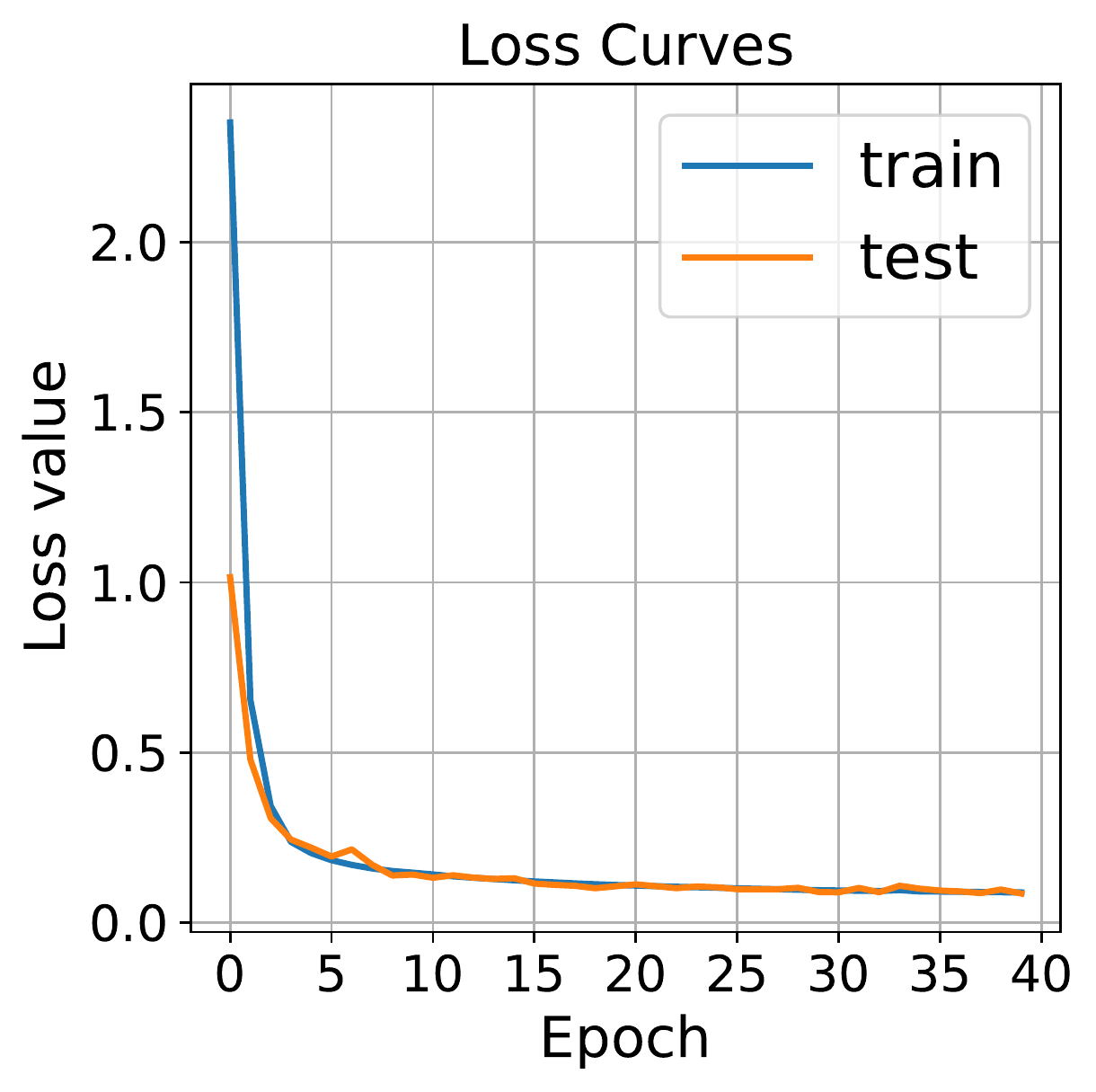} & \includegraphics[scale=0.2]{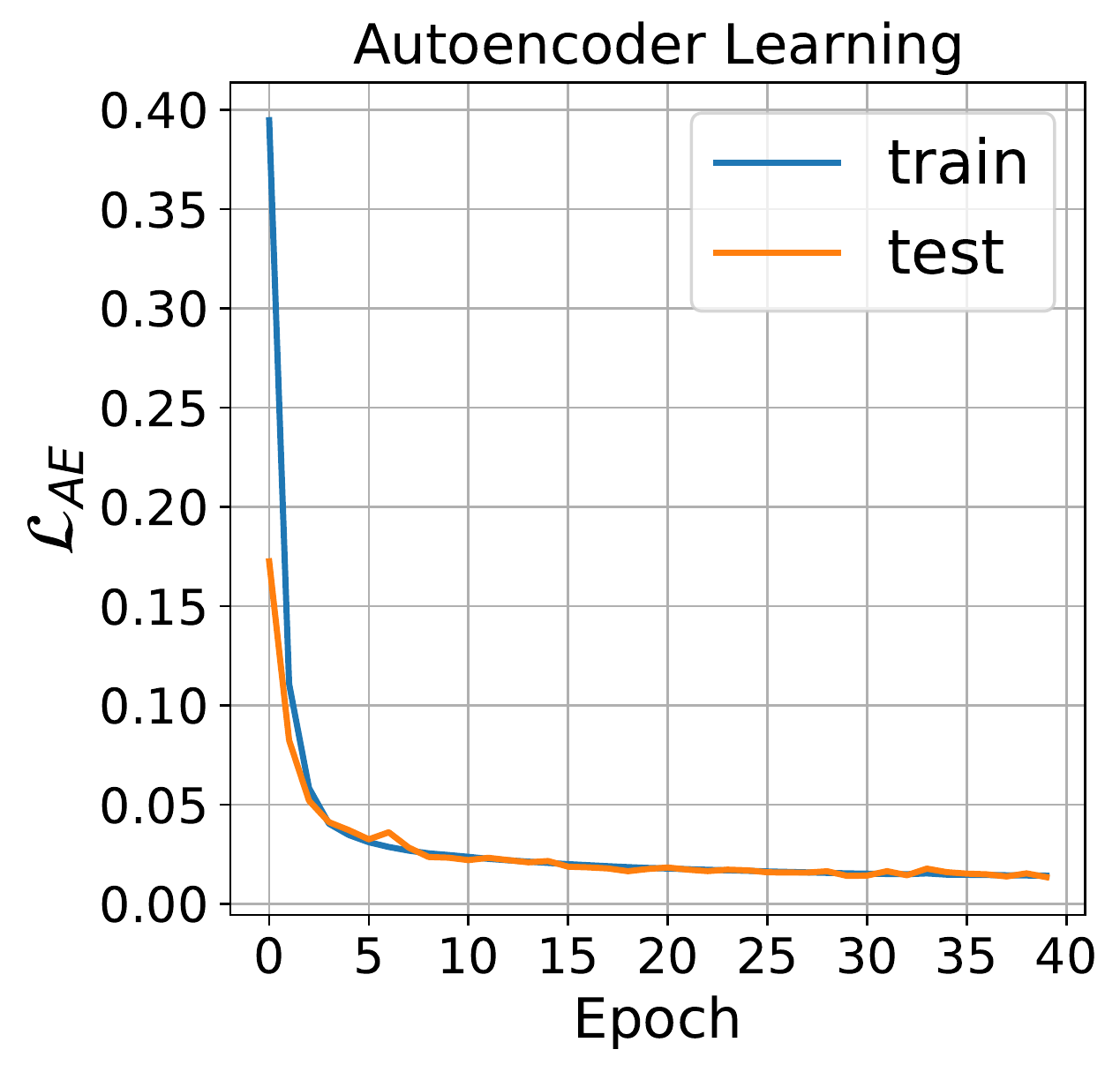}\\ 
                    \includegraphics[scale=0.2]{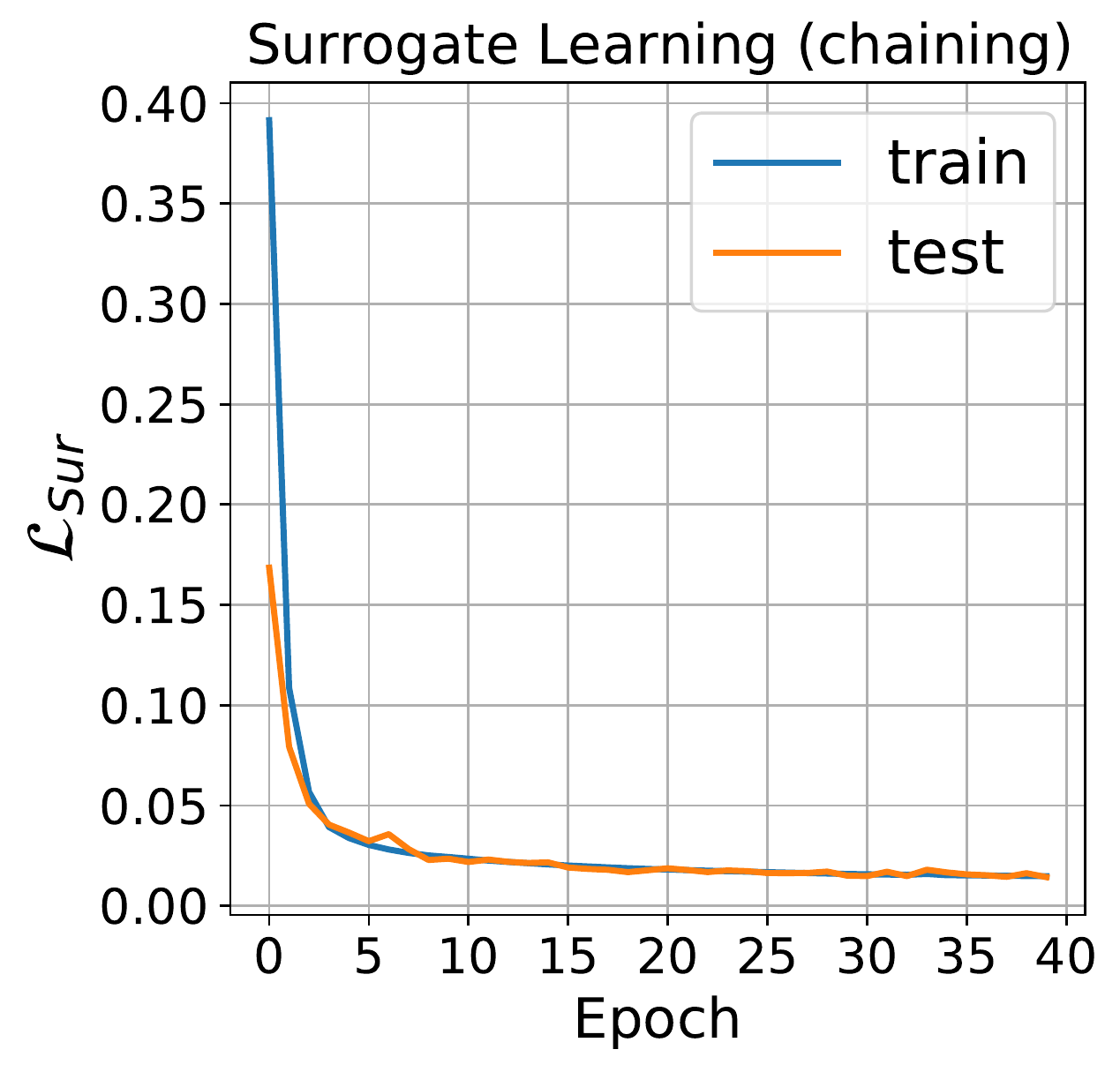} & \includegraphics[scale=0.2]{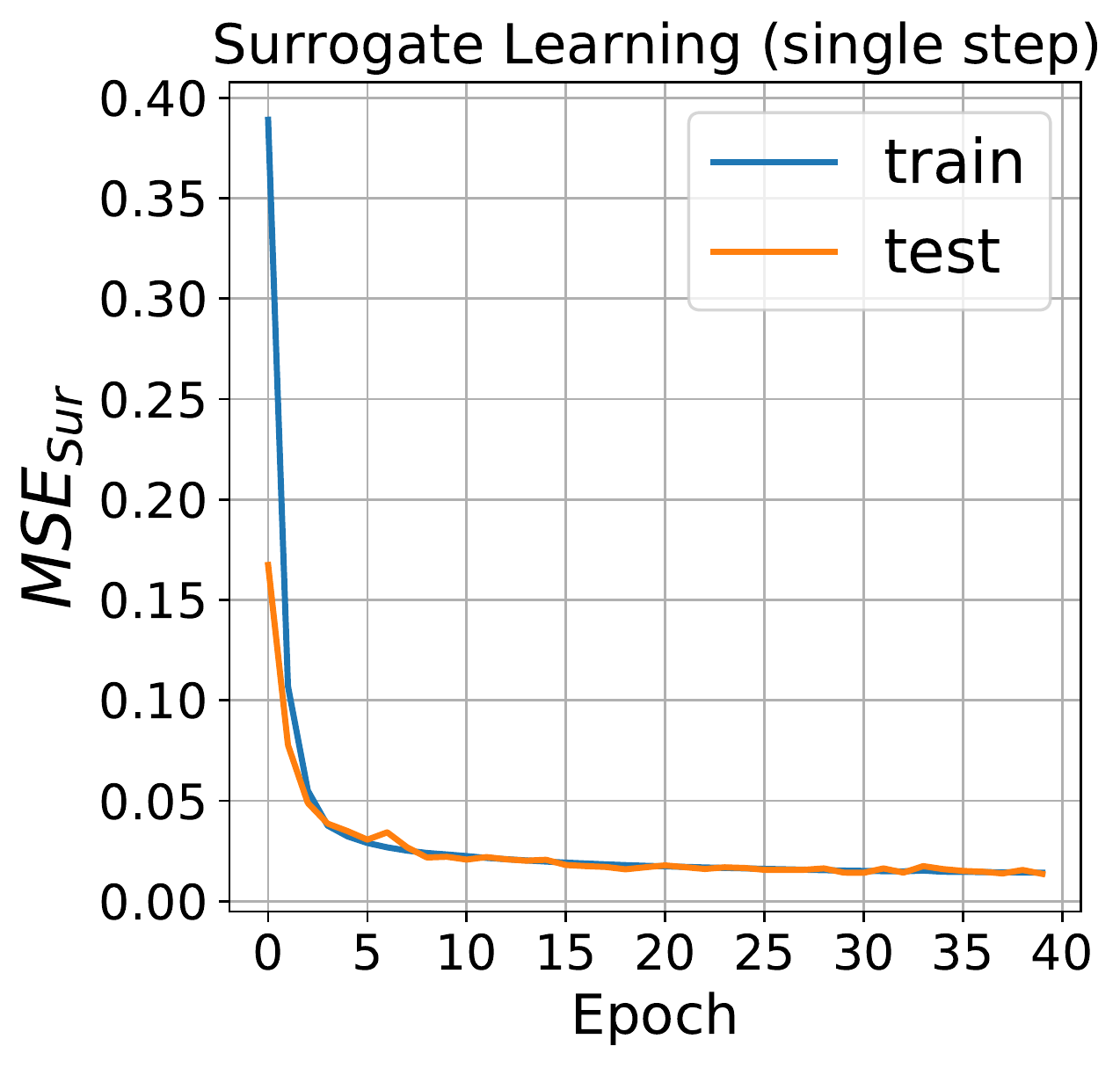}
                    \end{tabular}}
                    \caption{First plot (top left) represents the loss function (see \cref{eq:loss}) over learning. Second and third plots (top right and bottom left, respectively) separately show the two parts of \cref{eq:loss}. Last plot (bottom right) measures surrogate's efficiency but without chaining.}
                    \label{fig:losscurves}
        \end{figure}

        The training set represents $95\%$ of all available  data and the test set the remaining ones. The batch size is set  to $32$ and we chose an Adam optimizer with a learning rate of $10^{-3}$. The number of epochs is first arbitrarily fixed to $20$ for computation time purposes. Then, once a network shows satisfactory performances, it is retrained with $40$ epochs. Networks’ weights are saved each time we reach a lower loss score on the testset. Also, it is worth reminding that both the autoencoder and the surrogate are trained together: parameter $\rho$ of our custom loss function (see \cref{eq:loss}) is set to $5$ and the number of iterations $C$ is set to $2$. \Cref{fig:losscurves} consists of four curves that validate NNs effective learning: indeed, the smooth exponential decreasing of the loss function (see top left graph) indicates that NNs are performing the task they are assigned to better and better over training. Second and third plots (top right and bottom left, respectively) confirm that both the AE and the surrogate do learn, \textit{i.e.} that none of them is left behind during the training stage. Regarding the last plot (bottom right), it is close to the third one as it also measures the efficiency of the surrogate but without chaining, meaning that states are encoded, propagated only once and decoded afterwards. One could have noticed that $\mathcal{L}_{AE}$ and $\mathcal{L}_{Sur}$ have almost the same values over learning: it would suggest to set $\rho$ to $1$ rather than to $5$ in order to define a fair loss function. However, it turns out that weighting more $\mathcal{L}_{Sur}$ gives better scores, meaning that more effort is needed for the surrogate to properly learn than for the AE. Along with the aforementioned stability issues, it confirms that the surrogate network is the more challenging to train.\\
    
\begin{figure}[t]
            \centering
                \scalebox{0.25}{
                    \begin{tabular}{c c}
                    \includegraphics{original1.pdf} & \includegraphics{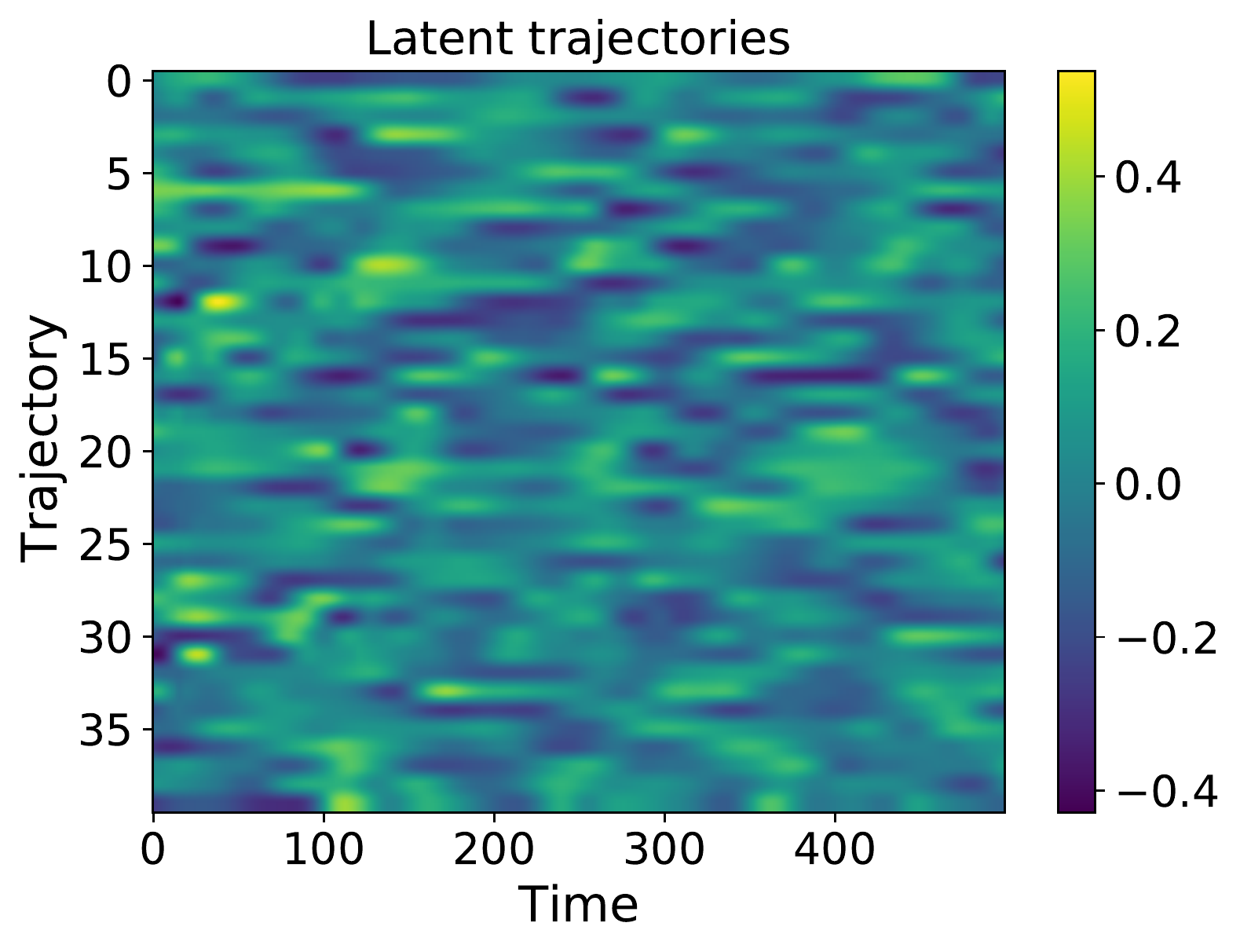}
                    
                    \end{tabular}}
                    \caption{First graph (left) represents the original Lorenz 96 dynamics in dimension $40$. Second graph (right) is the associated latent data yielded by the trained autoencoder.}
                    \label{fig:latentRecon}
        \end{figure}

        In \cref{sec:AE}, we pointed out the fact that the latent space produced by the autoencoder and the original $40$-variables Lorenz 96 dynamics have absolutely no reason to be alike: indeed, we remind that no constraint is added to the autoencoder in this sense. \Cref{fig:latentRecon} confirms this comment by putting aside both the Lorenz 96 data in dimension $40$ and the associated latent transformation: the autoencoder does not reproduce the original dynamics within its latent space. One could note that since the last activation function of the encoder is \textit{tanh} we could not expect the two plots to have the same order of magnitude and range of values.
        
    \subsection{Assessing the performance of our DA framework: the ETKF-Q-L}
    \label{sec:assessingPerf}

        We have proposed a new latent space DA algorithm coupled with an AE and a surrogate network. In this section we perform benchmark tests against the
        ETKF-Q-L algorithm presented in \cref{sec:latentETKF-Q}. For the comparison, the Root Mean Square Error (RMSE) is computed in the full space of dimension $400$:
        \begin{equation}
            \textit{RMSE}\left(\boldsymbol{x}_k,\mathcal{D}\left(\boldsymbol{z}_k\right)\right) = \sqrt{\frac{1}{\dimx}\sum_{i=1}^{\dimx} \left(\boldsymbol{x}_k^{(i)}-\mathcal{D}\left(\boldsymbol{z}_k^{(i)}\right)\right)^2}
        \label{eq:rmse}
        \end{equation}
        where $\mathcal{D}$ denotes the decoder, $\boldsymbol{z}_k$ is the latent prediction at time $k$ and $\boldsymbol{x}_k$ is the truth at the same time.\\
        
        Here is the list of the other approaches we benchmark our ETKF-Q-L method against:
        
        \begin{itemize}
        
            \item \textit{ETKF-Q}: we perform the standard ETKF-Q algorithm \cite{fillion} over the augmented Lorenz 96 data without resorting to any neural network. Propagation is performed by applying a standard Lorenz 96 propagator based on the Runge-Kutta fourth-order (RK4) scheme: this requires first to bring back the $400$-dimensional data to the $40$ dimensional space by applying transformation $\boldsymbol{\mathcal{F}}$ (see \cref{def:augmentedLorenz}). Data are mapped to the full space afterwards and the method loops back.
            
            \item \textit{ETKF-Q-P} (\textit{ETKF-Q-Physical}): it remains pretty close to the regular ETKF-Q algorithm (see \cref{algo:ETKF-Q}) as only the propagation step differs from it: instead of the standard propagator, we apply the encoder, the surrogate and the decoder, respectively. Notation \textit{-P} indicates that data assimilation is performed within the full - \textit{i.e.} physical - space.
            
            \item \textit{PCA-S-P} (\textit{Principal Component Analysis - Surrogate - Physical}): it works exactly like \textit{ETKF-Q-P} except that the encoder and the decoder are switched for a Principle Component Analysis (PCA) which can be seen as the simplest linear space reduction technique.
            
            \item \textit{PCA-S-L} (\textit{Principal Component Analysis - Surrogate - Latent}): this approach and ours are very alike, the only difference is that encoding and decoding stages are performed with a PCA. 
            
            \item \textit{PCA-LinReg-P} (\textit{Principal Component Analysis - Linear Regression - Physical}): same as \textit{PCA-S-P} but here the surrogate is replaced with the \textit{scikit-learn} linear regression predictor.
            
            \item \textit{PCA-LinReg-L} (\textit{Principal Component Analysis - Linear Regression - Latent}): same as \textit{PCA-S-L} but here again the surrogate is replaced with the \textit{scikit-learn} linear regression predictor.
        \end{itemize}
        
        \Cref{tab:res1} summarizes the benchmark context by specifying the DA space (\textit{i.e.} the space where analysis is performed) along with the propagation method used.\\
    
        In the standard ETKF algorithm, the multiplicative inflation parameter needs to be tuned. For simplicity, we chose $\boldsymbol{R}$ and $\boldsymbol{Q}$ as diagonal matrices: $\boldsymbol{R}=\boldsymbol{\sigma}_R^2 \boldsymbol{I}_p$ and $\boldsymbol{Q} = \boldsymbol{\sigma}_Q^2 \boldsymbol{I}_{\dimx}$.
        Then, in the ETKF-Q algorithm \cite{fillion}, in addition to the multiplicative inflation parameter, there is another parameter namely the standard deviation of the additive model error, $\boldsymbol{\sigma}_Q$. 
        The trajectories of the ensemble states are corrected using this model error at the beginning of each cycle (see \cref{algo:ETKF-Q} and \cref{algo:LatentETKF-Q}). These updated states are used as the forecast states. However, due to the sampling error, this model error at each step can be thought of as an additive type of inflation. Thus, this model error parameter may correct the sampling error as well. With this idea in mind, the values of the model error during propagation and the update of the deviation matrix may be different. According to our experiments, indeed taking different values gives better RMSE scores.
    
        As for the observation operator $\mathcal{H}$, it is set to $\boldsymbol{I}_p$, meaning that observation and state spaces are the same \textit{i.e.} $\dimx=p$.
        
        For the latent space algorithms, we introduce $\boldsymbol{Q}_{\nll} = \boldsymbol{\sigma}_{Q_{\nll}}^2 \boldsymbol{I}_{\nll}$. Since we do not know the propagation error of the surrogate model, we need to iteratively test dozens of $\boldsymbol{\sigma}_{Q_{\nll}}$ to find the one that best suits.\\
        
        After some hand-made experiments, we decided to find the best parameters combination in a straightforward manner through a grid search. The context of experiment is as follows:
        
        \begin{itemize}
            \item \underline{Parameters:}
            \begin{itemize}
                \item $\boldsymbol{40}$ ensemble members
                \item Observation error covariance matrix $\boldsymbol{\sigma_R}^2 \boldsymbol{I}_{400}$ with $\boldsymbol{\sigma_R}=\boldsymbol{1.0}$.
                \item Initial forecast error covariance matrix $\boldsymbol{\sigma_B}^2 \boldsymbol{I}_{400}$ at time $k=0$ with $\boldsymbol{\sigma_B}=\boldsymbol{0.3}$.
                \item $\boldsymbol{1000}$ iterations/time steps.
                \item Model error covariance matrix $\boldsymbol{\sigma_M}^2 \boldsymbol{I}_{40}$ (for \textit{ETKF-Q} method only): $\boldsymbol{\sigma_M}=\boldsymbol{0.3}$. 
            \end{itemize}
            
            \item \underline{Grid search:}
            \begin{itemize}
                \item \textit{inflation} ranges from $\boldsymbol{0.99}$ to $\boldsymbol{1.9}$.
                \item $\boldsymbol{\sigma}_Q/\boldsymbol{\sigma}_{Q_{\nll}}$ ranges from $\boldsymbol{10^{-7}}$ to $\boldsymbol{0.9}$.
            \end{itemize}
        \end{itemize}
        
        We conducted an experiment that consists in tuning \textit{inflation} and $\boldsymbol{\sigma}_Q$ or $\boldsymbol{\sigma}_{Q_{\nll}}$ parameters. Distinction between $\boldsymbol{\sigma}_Q$ and $\boldsymbol{\sigma}_{Q_{\nll}}$ only aims at differentiating full space data assimilation from latent space data assimilation, respectively.
        Results are given in \cref{tab:res2}.
        
        \begin{table}
            \begin{center}
                \hspace*{0cm}
                    \scalebox{0.6}{
                    \begin{tabular}{|c|c|c|c|}
                    \hline
                     Name & DA space & Propagation & Correction\\
                     \hline
                     \textit{ETKF-Q} & Full space & RK4 & $\boldsymbol{\sigma}_Q$\\
                     \hline
                     \textit{ETKF-Q-P} & Full space & ($\mathcal{D}$, $\mathcal{S}$, $\mathcal{E}$) & $\boldsymbol{\sigma}_Q$\\
                     \hline
                     \textit{ETKF-Q-L} & Latent space & $\mathcal{S}$ & $\boldsymbol{\sigma}_{Q_{\nll}}$\\
                     \hline
                     \textit{PCA-S-P} & Full space & (\textit{PCA}, $\mathcal{S}$, \textit{reverse PCA}) & $\boldsymbol{\sigma}_Q$\\
                     \hline
                     \textit{PCA-S-L} & Latent space & $\mathcal{S}$ & $\boldsymbol{\sigma}_{Q_{\nll}}$\\
                     \hline
                     \textit{PCA-LinReg-P} & Full space & (\textit{PCA}, \textit{LinReg}, \textit{reverse PCA}) & $\boldsymbol{\sigma}_Q$\\
                     \hline
                     \textit{PCA-LinReg-L} & Latent space & \textit{LinReg} & $\boldsymbol{\sigma}_{Q_{\nll}}$\\
                     \hline
                 \end{tabular}
                 }
            \caption{Data Assimilation benchmark context. DA space denotes the analysis space.} 
            \label{tab:res1}
            \end{center}
            
        \end{table}

        \begin{table}
            \begin{center}
                \hspace*{0cm}
                    \scalebox{0.8}{
                    \begin{tabular}{|c|c|c|c|}
                    \hline
                     Name & RMSE & Inflation & $\boldsymbol{\sigma}_Q/\boldsymbol{\sigma}_{Q_{\nll}}$\\
                     \hline
                     \textit{ETKF-Q} & \textcolor{red}{$0.194$} & $1.12$ & $0.07$ \\
                     \hline
                     \textit{ETKF-Q-P} & \textcolor{red}{$0.217$} & $1.08$ & $0.1$\\
                     \hline
                     \textbf{\textit{ETKF-Q-L}} & \textcolor{red}{$\boldsymbol{0.168}$} & $\boldsymbol{1.004}$ & $\boldsymbol{5.10^{-5}}$\\
                     \hline
                     \textit{PCA-S-P} & \textcolor{red}{$0.383$} & $1.145$ & $0.5$\\
                     \hline
                     \textit{PCA-S-L} & \textcolor{red}{$0.383$} & $1.13$ & $0.6$\\
                     \hline
                     \textit{PCA-LinReg-P} & \textcolor{red}{$0.429$} & $1.24$ & $0.7$\\
                     \hline
                     \textit{PCA-LinReg-L} & \textcolor{red}{$0.426$} & $1.2$ & $0.9$\\
                     \hline
                 \end{tabular}
                 }
            \caption{Data Assimilation results (mean RMSE). Only two parameters are tuned: \textit{inflation} and $\boldsymbol{\sigma}_Q/\boldsymbol{\sigma}_{Q_{\nll}}$.} 
            \label{tab:res2}
            \end{center}
            
        \end{table}
        
        They draw the conclusion that our approach is competitive with other methods that rely on a simple space reduction technique like PCA or on a very straightforward propagator such as the linear predictor from the \textit{scikit-learn} library. As best RMSE score is reached for our solution, it suggests that the autoencoder fully leverages nonlinear transformations to compress the information hold by the augmented Lorenz 96 dynamics (see \cref{fig:illustration}). Besides, it proves that for this particular tailored system, a mere Multi-Layer Perceptron (MLP) can properly perform the time propagation step in a suitable latent space: more precisely, the experiments do not show that a simple MLP can propagate through time, but it rather demonstrates that surrounded with an encoder and a decoder, a MLP can replace the model. Our training probably finds a latent space where the MLP surrogate can perform well the time propagation. Although the proposed approach seems both simple and efficient, we remind that it was nonetheless not straightforward and we needed to resort to a custom loss function with an iterative training scheme that enhances stability in order to achieve a satisfactory solving.\\
        
        As priorly mentioned in the introduction, our method also aims at tackling time computing issues faced by regular DA algorithms. Therefore, all our benchmark algorithms have been run $100$ times on a virtual machine composed of four virtual CPUs, one NVIDIA Tesla P100 GPU and $15$ GB of RAM. \Cref{tab:CompTime} evidences that our method computes $2.4$ to $2.5$ times faster than the standard ETKF-Q algorithm. More generally, we notice that as long as an algorithm utilizes a latent space structure to perform DA analysis, a significant computational gain is obtained. We do not observe a major difference between GPU and CPU computations except for \textit{ETKF-Q-P} algorithm. Yet, standard deviations seem to be larger when using CPU.
        Crossing these results with \cref{tab:res2} clearly reveals that our approach is the best one in our benchmark context on both the accuracy and the computational cost criteria.

\begin{table}

        \begin{center}

            \hspace*{-0.2cm}

                \scalebox{0.85}{

                \begin{tabular}{|c|c||c|c||c|}

                \hline

                \multirow{2}{3em}{Name} & \multicolumn{2}{|c|}{GPU} & \multicolumn{2}{|c|}{CPU} \\

                \cline{2-5}

                & Avg. Time & Std & Avg. Time & Std\\

                 \hline

                 \textit{ETKF-Q} & \textcolor{red}{$16.32s$} & \textcolor{red}{$0.20s$} & \textcolor{red}{$16.23s$} & \textcolor{red}{$0.31s$}\\

                 \hline

                 \textit{ETKF-Q-P} & \textcolor{red}{$13.60s$} & \textcolor{red}{$0.19s$} & \textcolor{red}{$17.12s$} & \textcolor{red}{$0.84s$}\\

                 \hline

                 \textbf{\textit{ETKF-Q-L}} & \textcolor{red}{$\boldsymbol{6.52s}$} & \textcolor{red}{$\boldsymbol{0.12s}$} & \textcolor{red}{$\boldsymbol{6.89s}$} & \textcolor{red}{$\boldsymbol{0.62s}$}\\

                 \hline

                 \textit{PCA-S-P} & \textcolor{red}{$13.22s$} & \textcolor{red}{$0.18s$} & \textcolor{red}{$12.18s$} & \textcolor{red}{$0.31s$}\\

                 \hline

                 \textbf{\textit{PCA-S-L}} & \textcolor{red}{$\boldsymbol{5.93s}$} & \textcolor{red}{$\boldsymbol{0.10s}$} & \textcolor{red}{$\boldsymbol{5.35s}$} & \textcolor{red}{$\boldsymbol{0.16s}$}\\

                 \hline

                 \textit{PCA-LinReg-P} & \textcolor{red}{$12.09s$} & \textcolor{red}{$0.29s$} & \textcolor{red}{$11.62s$} & \textcolor{red}{$0.27s$}\\

                 \hline

                 \textbf{\textit{PCA-LinReg-L}} & \textcolor{red}{$\boldsymbol{5.02s}$} & \textcolor{red}{$\boldsymbol{0.12s}$} & \textcolor{red}{$\boldsymbol{4.94s}$} & \textcolor{red}{$\boldsymbol{0.15s}$}\\

                 \hline

             \end{tabular}

             }

        \caption{GPU and CPU average computational times over $100$ runs for all the algorithms tested in this case study. Resorting to a GPU device is only meaningful for methods that include NNs. Std denotes the standard deviation.}

        \label{tab:CompTime}

        \end{center}
            
        \end{table}

\section{Conclusion and Outlook}

    Broadening DL algorithms to solve physics is justified by its efficiency in space reduction and time propagation tasks. Whereas prior aims like performing image classification, segmentation or language processing have been addressed by NNs solely, today's challenges involve using DL in more global processes. In this paper, we proposed to study how to incorporate DL in a DA framework. For this proof of concept, we supposed that we were given a dynamical system lying in $\mathbb{R}^{\dimx}$ exhibiting a latent representation of lower dimension $\nll$. Under this assumption, we proved that it is possible to perform the ETKF-Q algorithm within a latent space of same size $\nll$ produced by an autoencoder. While performing the ETKF-Q algorithm in a latent space, we take proper account of the necessary model calibration needed in the evolution model description in latent space.
  
    The motivation for performing the latent space DA was both reducing the computational cost and also getting a better accuracy. Reducing the computational cost is a natural gain due to performing DA in a reduced space. The accuracy gain on the other hand depends on the fact that 
    DA linear analysis in the latent space obtained by the AE is less susceptible to yield non-physical solutions. 
    
    We have therefore trained an AE and a surrogate network through a single learning thanks to a chained custom loss function. In addition to allowing a better training since both the AE and the surrogate modify their weights according to each other, this particular loss function also enhances stability. Indeed, the surrogate is called several times in a row and thus has to produce stable results at least for two successive time steps. Experience demonstrates that this is already enough to get a quite satisfactory stable behaviour on longer time windows. Then, given these two networks we can perform latent DA.
    
    We have shown the potential of our methodology on the instructional augmented Lorenz 96 system which is designed such that we ensure the existence of a latent dynamics. We have compared our methodology to the existing ETKF-Q algorithm and numerical results confirmed that the proposed methodology performs better than the usual full space strategy. Our methodology can be utilized as long as a system lying in $\mathbb{R}^{\dimx}$ is accurately representable in a lower dimension.

    We believe that the proposed proof of concept is encouraging and that, as such, it stimulates several tracks for future research. For instance, to before considering a use in more operational situations, a number of theoretical question should be considered. For instance, it would be interesting to further investigate the sensibility of the methodology to several hyper-parameters of the method including the latent space dimension, observation frequency, inflation etc. It is also important to better understand  how the properties of the full model dynamics behave in the latent space including the model error. 
    We remind that in this study we consider very simple NN architectures. Investigating more sophisticated networks and introducing constraints to enforce latent space's structure could be of course important. This structure may be motivated by underlying properties of the physics, such as the coupling between variables, or other structures. This being done, considering other toy problems would necessary before going for more complex problems.
    
    Finally, the latent space strategy that we propose in this paper for an ensemble Kalman filter type method is quite general in the sense that it can be easily adapted to other DA algorithms like the variational approaches. 

\section*{Acknowledgements}

    We are very thankful to Atos, ANITI, CERFACS, University of Toulouse, and NVIDIA for supporting and being involved in Mathis Peyron's PhD. They have provided us with computational resources, materials, funds and experts' advice that have been core in this research.


\begin{thebibliography}{10}\itemsep=-1pt

\bibitem{Amezcua2017}
J.~Amezcua, M.~Goodliff, and P.~J.~V. Leeuwen.
\newblock A weak-constraint 4densemblevar. part i: formulation and simple model
  experiments.
\newblock {\em Tellus A: Dynamic Meteorology and Oceanography}, 69(1):1271564,
  2017.

\bibitem{artana}
G.~Artana, A.~Cammilleri, J.~Carlier, and E.~M{\'e}min.
\newblock {Strong and weak constraint variational assimilations for reduced
  order fluid flow modeling}.
\newblock {\em {Journal of Computational Physics}}, 213(8):3264--3288, Apr.
  2012.

\bibitem{BocquetSiam}
M.~Asch, M.~Bocquet, and M.~Nodet.
\newblock {\em {Data assimilation: methods, algorithms, and applications}},
  chapter 1: Introduction to data assimilation and inverse problems, pages
  3--23.
\newblock Fundamentals of Algorithms. {SIAM}, 2016.

\bibitem{BocquetSiamTot}
M.~Asch, M.~Bocquet, and M.~Nodet.
\newblock {\em {Data assimilation: methods, algorithms, and applications}}.
\newblock Fundamentals of Algorithms. {SIAM}, 2016.

\bibitem{bachlechner2020rezero}
T.~Bachlechner, B.~P. Majumder, H.~H. Mao, G.~W. Cottrell, and J.~McAuley.
\newblock Rezero is all you need: Fast convergence at large depth, 2020.

\bibitem{Bannister2008}
R.~N. Bannister.
\newblock A review of forecast error covariance statistics in atmospheric
  variational data assimilation. i: Characteristics and measurements of
  forecast error covariances.
\newblock {\em Quarterly Journal of the Royal Meteorological Society},
  134(637):1951--1970, 2008.

\bibitem{BocquetLecture}
M.~Bocquet.
\newblock Lecture notes, 2014, last revision: January 2019.

\bibitem{bocquet:hal-01592362}
M.~Bocquet and A.~Carrassi.
\newblock {Four-dimensional ensemble variational data assimilation and the
  unstable subspace}.
\newblock {\em {Tellus A}}, 69(1):1304504, Mar. 2017.

\bibitem{Bocquet2014}
M.~Bocquet and P.~Sakov.
\newblock An iterative ensemble kalman smoother.
\newblock {\em Quarterly Journal of the Royal Meteorological Society},
  140(682):1521--1535, 2014.

\bibitem{Boukabara2020}
S.-A. Boukabara, V.~Krasnopolsky, S.~G. Penny, J.~Q. Stewart, A.~McGovern,
  D.~Hall, J.~E.~T. Hoeve, J.~Hickey, H.-L.~A. Huang, J.~K. Williams, K.~Ide,
  P.~Tissot, S.~E. Haupt, K.~S. Casey, N.~Oza, A.~J. Geer, E.~S. Maddy, and
  R.~N. Hoffman.
\newblock Outlook for exploiting artificial intelligence in the earth and
  environmental sciences.
\newblock {\em Bulletin of the American Meteorological Society}, pages 1 -- 53,
  20 Nov. 2020.

\bibitem{Brajard_2020}
J.~Brajard, A.~Carrassi, M.~Bocquet, and L.~Bertino.
\newblock Combining data assimilation and machine learning to emulate a
  dynamical model from sparse and noisy observations: A case study with the
  lorenz 96 model.
\newblock {\em Journal of Computational Science}, 44:101171, Jul 2020.

\bibitem{Canchumuni2019}
S.~W. Canchumuni, A.~A. Emerick, and M.~A.~C. Pacheco.
\newblock Towards a robust parameterization for conditioning facies models
  using deep variational autoencoders and ensemble smoother.
\newblock {\em Computers \& Geosciences}, 128:87–102, Jul 2019.

\bibitem{Cao2006b}
Y.~Cao, J.~Zhu, I.~M. Navon, and Z.~Luo.
\newblock A reduced-order approach to four-dimensional variational data
  assimilation using proper orthogonal decomposition.
\newblock {\em International Journal for Numerical Methods in Fluids},
  53(10):1571--1583, 2007.

\bibitem{CarliniWagner}
N.~Carlini and D.~A. Wagner.
\newblock Towards evaluating the robustness of neural networks.
\newblock {\em CoRR}, abs/1608.04644, 2016.

\bibitem{CarrassiTrevisan2008}
A.~Carrassi, A.~Trevisan, L.~Descamps, O.~Talagrand, and F.~Uboldi.
\newblock Controlling instabilities along a 3dvar analysis cycle by
  assimilating in the unstable subspace: a comparison with the enkf.
\newblock {\em Nonlinear Processes in Geophysics}, 15(4):503--521, 2008.

\bibitem{CarrassiTrevisan2007}
A.~Carrassi, A.~Trevisan, and F.~Uboldi.
\newblock Adaptive observations and assimilation in the unstable subspace by
  breeding on the data-assimilation system.
\newblock {\em Tellus A}, 59(1):101--113, 2007.

\bibitem{Cohn1996}
S.~E. Cohn and R.~Todling.
\newblock Approximate data assimilation schemes for stable and unstable
  dynamics.
\newblock {\em Journal of the Meteorological Society of Japan. Ser. II},
  74(1):63--75, 1996.

\bibitem{fillion}
A.~Fillion, M.~Bocquet, S.~Gratton, S.~Gürol, and P.~Sakov.
\newblock An iterative ensemble kalman smoother in presence of additive model
  error.
\newblock {\em SIAM/ASA Journal on Uncertainty Quantification}, 8(1):198--228,
  2020.

\bibitem{FultonDeformable2018}
L.~Fulton, V.~Modi, D.~Duvenaud, D.~I.~W. Levin, and A.~Jacobson.
\newblock Latent-space dynamics for reduced deformable simulation.
\newblock {\em Computer Graphics Forum}, 2019.

\bibitem{garcia2017review}
A.~Garcia-Garcia, S.~Orts-Escolano, S.~Oprea, V.~Villena-Martinez,
  P.~Martinez-Gonzalez, and J.~Garcia-Rodriguez.
\newblock A survey on deep learning techniques for image and video semantic
  segmentation.
\newblock {\em Applied Soft Computing}, 70:41--65, 2018.

\bibitem{gardner2018allennlp}
M.~Gardner, J.~Grus, M.~Neumann, O.~Tafjord, P.~Dasigi, N.~F. Liu, M.~Peters,
  M.~Schmitz, and L.~Zettlemoyer.
\newblock {A}llen{NLP}: A deep semantic natural language processing platform.
\newblock In {\em Proceedings of Workshop for {NLP} Open Source Software
  ({NLP}-{OSS})}, pages 1--6, Melbourne, Australia, July 2018. Association for
  Computational Linguistics.

\bibitem{ECMWFarticle}
A.~J. Geer.
\newblock Learning earth system models from observations: machine learning or
  data assimilation?
\newblock {\em Philosophical Transactions of the Royal Society A: Mathematical,
  Physical and Engineering Sciences}, 379(2194):20200089, 2021.

\bibitem{haber2019imexnet}
E.~Haber, K.~Lensink, E.~Treister, and L.~Ruthotto.
\newblock Imexnet: A forward stable deep neural network, 2019.

\bibitem{Haber_2017}
E.~Haber and L.~Ruthotto.
\newblock Stable architectures for deep neural networks.
\newblock {\em Inverse Problems}, 34(1):014004, dec 2017.

\bibitem{Han_2018}
J.~Han, A.~Jentzen, and W.~E.
\newblock Solving high-dimensional partial differential equations using deep
  learning.
\newblock {\em Proceedings of the National Academy of Sciences},
  115(34):8505–8510, Aug 2018.

\bibitem{he2015deep}
K.~He, X.~Zhang, S.~Ren, and J.~Sun.
\newblock Deep residual learning for image recognition.
\newblock {\em 2016 IEEE Conference on Computer Vision and Pattern Recognition
  (CVPR)}, pages 770--778, 2016.

\bibitem{Khoo2018SolvingPP}
Y.~Khoo, J.~Lu, and L.~Ying.
\newblock Solving parametric pde problems with artificial neural networks.
\newblock {\em European Journal of Applied Mathematics}, page 1–15, 2020.

\bibitem{Kramer1991}
M.~A. Kramer.
\newblock Nonlinear principal component analysis using autoassociative neural
  networks.
\newblock {\em AIChE Journal}, 37(2):233--243, 1991.

\bibitem{Krizhevsky}
A.~Krizhevsky, I.~Sutskever, and G.~E. Hinton.
\newblock Imagenet classification with deep convolutional neural networks.
\newblock In F.~Pereira, C.~J.~C. Burges, L.~Bottou, and K.~Q. Weinberger,
  editors, {\em Advances in Neural Information Processing Systems}, volume~25,
  pages 1097--1105. Curran Associates, Inc., 2012.

\bibitem{Kuenzer}
C.~Kuenzer, M.~Ottinger, M.~Wegmann, H.~Guo, C.~Wang, J.~Zhang, S.~Dech, and
  M.~Wikelski.
\newblock Earth observation satellite sensors for biodiversity monitoring:
  potentials and bottlenecks.
\newblock {\em International Journal of Remote Sensing}, 35(18):6599--6647,
  2014.

\bibitem{le2012building}
Q.~V. Le, M.~Ranzato, R.~Monga, M.~Devin, K.~Chen, G.~S. Corrado, J.~Dean, and
  A.~Y. Ng.
\newblock Building high-level features using large scale unsupervised learning.
\newblock In {\em Proceedings of the 29th International Coference on
  International Conference on Machine Learning}, ICML'12, page 507–514,
  Madison, WI, USA, 2012. Omnipress.

\bibitem{li2020fourier}
Z.~{Li}, N.~{Kovachki}, K.~{Azizzadenesheli}, B.~{Liu}, K.~{Bhattacharya},
  A.~{Stuart}, and A.~{Anandkumar}.
\newblock {Fourier Neural Operator for Parametric Partial Differential
  Equations}.
\newblock {\em arXiv e-prints}, page arXiv:2010.08895, Oct. 2020.

\bibitem{Lorenz96}
E.~Lorenz.
\newblock Predictability: a problem partly solved.
\newblock In {\em Seminar on Predictability, 4-8 September 1995}, volume~1,
  pages 1--18, Shinfield Park, Reading, 1995. ECMWF, ECMWF.

\bibitem{Lu2007SequentialDA}
Z.~Lu, T.~K. Leen, R.~van~der Merwe, S.~Frolov, and A.~M. Baptista.
\newblock Sequential data assimilation with sigma-point kalman filter on
  low-dimensional manifold.
\newblock {\em submitted to Journal of Inverse Problems}, 2007.

\bibitem{Mack2020AttentionbasedCA}
J.~Mack, R.~Arcucci, M.~Molina-Solana, and Y.~Guo.
\newblock Attention-based convolutional autoencoders for 3d-variational data
  assimilation.
\newblock {\em Computer Methods in Applied Mechanics and Engineering},
  372:113291, 2020.

\bibitem{Mandel2016}
J.~Mandel, E.~Bergou, S.~G\"urol, S.~Gratton, and I.~Kasanický.
\newblock Hybrid levenberg--marquardt and weak-constraint ensemble kalman
  smoother~method.
\newblock {\em Nonlinear Processes in Geophysics}, 23(2):59--73, 2016.

\bibitem{maulik2020reducedorder}
R.~Maulik, B.~Lusch, and P.~Balaprakash.
\newblock Reduced-order modeling of advection-dominated systems with recurrent
  neural networks and convolutional autoencoders.
\newblock {\em Physics of Fluids}, 33(3):037106, 2021.

\bibitem{Miller1989}
R.~N. Miller and M.~A. Cane.
\newblock A kalman filter analysis of sea level height in the tropical pacific.
\newblock {\em Journal of Physical Oceanography}, 19(6):773 -- 790, 01 Jun.
  1989.

\bibitem{Mitchell2015}
L.~Mitchell and A.~Carrassi.
\newblock Accounting for model error due to unresolved scales within ensemble
  kalman filtering.
\newblock {\em Quarterly Journal of the Royal Meteorological Society},
  141(689):1417--1428, 2015.

\bibitem{mohan2020embedding}
A.~Mohan, N.~Lubbers, D.~Livescu, and M.~Chertkov.
\newblock Embedding hard physical constraints in neural network coarse-graining
  of 3d turbulence.
\newblock {\em arXiv: Computational Physics}, 2020.

\bibitem{otter2019survey}
D.~W. Otter, J.~R. Medina, and J.~K. Kalita.
\newblock A survey of the usages of deep learning for natural language
  processing.
\newblock {\em IEEE Transactions on Neural Networks and Learning Systems},
  2020.

\bibitem{pawar2020data}
S.~M. Pawar and O.~San.
\newblock Data assimilation empowered neural network parameterizations for
  subgrid processes in geophysical flows.
\newblock {\em arXiv: Computational Physics}, 2020.

\bibitem{qi2017pointnet}
C.~Qi, H.~Su, K.~Mo, and L.~Guibas.
\newblock Pointnet: Deep learning on point sets for 3d classification and
  segmentation.
\newblock {\em 2017 IEEE Conference on Computer Vision and Pattern Recognition
  (CVPR)}, pages 77--85, 2017.

\bibitem{raissi2017physics1}
M.~Raissi, P.~Perdikaris, and G.~Karniadakis.
\newblock Physics informed deep learning (part i): Data-driven solutions of
  nonlinear partial differential equations.
\newblock {\em ArXiv}, abs/1711.10561, 2017.

\bibitem{raissi2017physics2}
M.~Raissi, P.~Perdikaris, and G.~Karniadakis.
\newblock Physics informed deep learning (part ii): Data-driven discovery of
  nonlinear partial differential equations.
\newblock {\em ArXiv}, abs/1711.10566, 2017.

\bibitem{Raissi_2018}
M.~Raissi, Z.~Wang, M.~S. Triantafyllou, and G.~E. Karniadakis.
\newblock Deep learning of vortex-induced vibrations.
\newblock {\em Journal of Fluid Mechanics}, 861:119–137, Dec 2018.

\bibitem{Reichstein2019DeepLA}
M.~Reichstein, G.~Camps-Valls, B.~Stevens, M.~Jung, J.~Denzler, N.~Carvalhais,
  and Prabhat.
\newblock Deep learning and process understanding for data-driven earth system
  science.
\newblock {\em Nature}, 566:195--204, 2019.

\bibitem{rottmann2020detection}
M.~Rottmann, K.~Maag, M.~Peyron, N.~Krejic, and H.~Gottschalk.
\newblock Detection of iterative adversarial attacks via counter attack, 2021.

\bibitem{SakovAsynch}
P.~Sakov and M.~Bocquet.
\newblock Asynchronous data assimilation with the enkf in presence of additive
  model error.
\newblock {\em Tellus A: Dynamic Meteorology and Oceanography}, 70(1):1--7,
  2018.

\bibitem{Sakov2018}
P.~Sakov, J.-M. Haussaire, and M.~Bocquet.
\newblock An iterative ensemble kalman filter in the presence of additive model
  error.
\newblock {\em Quarterly Journal of the Royal Meteorological Society},
  144(713):1297--1309, 2018.

\bibitem{seo2019graphs}
S.~Seo and Y.~Liu.
\newblock Differentiable physics-informed graph networks.
\newblock {\em ArXiv}, abs/1902.02950, 2019.

\bibitem{silver2016mastering}
D.~Silver, A.~Huang, C.~J. Maddison, A.~Guez, L.~Sifre, G.~Van Den~Driessche,
  J.~Schrittwieser, I.~Antonoglou, V.~Panneershelvam, M.~Lanctot, et~al.
\newblock Mastering the game of go with deep neural networks and tree search.
\newblock {\em nature}, 529(7587):484--489, 2016.

\bibitem{silver2017mastering}
D.~Silver, J.~Schrittwieser, K.~Simonyan, I.~Antonoglou, A.~Huang, A.~Guez,
  T.~Hubert, L.~Baker, M.~Lai, A.~Bolton, et~al.
\newblock Mastering the game of go without human knowledge.
\newblock {\em nature}, 550(7676):354--359, 2017.

\bibitem{simonyan2015deep}
K.~Simonyan and A.~Zisserman.
\newblock Very deep convolutional networks for large-scale image recognition.
\newblock {\em CoRR}, abs/1409.1556, 2015.

\bibitem{sutskever2014sequence}
I.~Sutskever, O.~Vinyals, and Q.~V. Le.
\newblock Sequence to sequence learning with neural networks.
\newblock In {\em Proceedings of the 27th International Conference on Neural
  Information Processing Systems - Volume 2}, NIPS'14, page 3104–3112,
  Cambridge, MA, USA, 2014. MIT Press.

\bibitem{Szegedy2013IntriguingPO}
C.~Szegedy, W.~Zaremba, I.~Sutskever, J.~Bruna, D.~Erhan, I.~J. Goodfellow, and
  R.~Fergus.
\newblock Intriguing properties of neural networks.
\newblock {\em CoRR}, abs/1312.6199, 2013.

\bibitem{TANGBORN2000}
A.~Tangborn and S.~Q. Zhang.
\newblock Wavelet transform adapted to an approximate kalman filter system.
\newblock {\em Applied Numerical Mathematics}, 33(1):307--316, 2000.

\bibitem{Trevisan2010}
A.~Trevisan, M.~D'Isidoro, and O.~Talagrand.
\newblock Four-dimensional variational assimilation in the unstable subspace
  and the optimal subspace dimension.
\newblock {\em Quarterly Journal of the Royal Meteorological Society},
  136(647):487--496, 2010.

\bibitem{Trevisan2011}
A.~Trevisan and L.~Palatella.
\newblock On the kalman filter error covariance collapse into the unstable
  subspace.
\newblock {\em Nonlinear Processes in Geophysics}, 18(2):243--250, 2011.

\bibitem{TrevisanUboldi2004}
A.~Trevisan and F.~Uboldi.
\newblock Assimilation of standard and targeted observations within the
  unstable subspace of the observation–analysis–forecast cycle system.
\newblock {\em Journal of the Atmospheric Sciences}, 61(1):103 -- 113, 01 Jan.
  2004.

\bibitem{TUANPHAM1998}
D.~{Tuan Pham}, J.~Verron, and M.~{Christine Roubaud}.
\newblock A singular evolutive extended kalman filter for data assimilation in
  oceanography.
\newblock {\em Journal of Marine Systems}, 16(3):323--340, 1998.

\bibitem{UboldiTrevisan2006}
F.~Uboldi and A.~Trevisan.
\newblock Detecting unstable structures and controlling error growth by
  assimilation of standard and adaptive observations in a primitive equation
  ocean model.
\newblock {\em Nonlinear Processes in Geophysics}, 13(1):67--81, 2006.

\bibitem{Vlachas2018}
P.~R. Vlachas, W.~Byeon, Z.~Y. Wan, T.~P. Sapsis, and P.~Koumoutsakos.
\newblock Data-driven forecasting of high-dimensional chaotic systems with long
  short-term memory networks.
\newblock {\em Proceedings of the Royal Society A: Mathematical, Physical and
  Engineering Sciences}, 474(2213):20170844, May 2018.

\bibitem{vonrueden2020informed}
L.~von R{\"u}den, S.~Mayer, K.~Beckh, B.~Georgiev, S.~Giesselbach, R.~Heese,
  B.~Kirsch, J.~Pfrommer, A.~Pick, R.~Ramamurthy, M.~Walczak, J.~Garcke,
  C.~Bauckhage, and J.~Sch{\"u}cker.
\newblock Informed machine learning -- a taxonomy and survey of integrating
  knowledge into learning systems.
\newblock {\em arXiv: Machine Learning}, 2019.

\bibitem{Wiewel2018}
S.~Wiewel, M.~Becher, and N.~Thuerey.
\newblock Latent-space physics: Towards learning the temporal evolution of
  fluid flow.
\newblock {\em CoRR}, abs/1802.10123, 2018.

\bibitem{Wiewel2020}
S.~Wiewel, B.~Kim, V.~C. Azevedo, B.~Solenthaler, and N.~Thuerey.
\newblock Latent space subdivision: Stable and controllable time predictions
  for fluid flow.
\newblock {\em Computer Graphics Forum}, 39(8):15--25, 2020.

\bibitem{wu2016googles}
Y.~Wu, M.~Schuster, Z.~Chen, Q.~V. Le, M.~Norouzi, W.~Macherey, M.~Krikun,
  Y.~Cao, Q.~Gao, K.~Macherey, J.~Klingner, A.~Shah, M.~Johnson, X.~Liu,
  L.~Kaiser, S.~Gouws, Y.~Kato, T.~Kudo, H.~Kazawa, K.~Stevens, G.~Kurian,
  N.~Patil, W.~Wang, C.~Young, J.~Smith, J.~Riesa, A.~Rudnick, O.~Vinyals,
  G.~Corrado, M.~Hughes, and J.~Dean.
\newblock Google's neural machine translation system: Bridging the gap between
  human and machine translation.
\newblock {\em CoRR}, abs/1609.08144, 2016.

\end{thebibliography}
\end{document}